\documentclass[journal]{IEEEtran}

\usepackage{amsfonts}
\usepackage{balance}
\usepackage{booktabs}
\usepackage{amsmath}
\usepackage{tabularx}
\usepackage{subfigure}
\usepackage{graphicx}
\usepackage{multirow}
\usepackage{rotating}
\usepackage{algorithmic}
\usepackage{footnote}
\usepackage{threeparttable}
\usepackage{times}
\usepackage{color}

\begin{document}

\title{Enhanced Ensemble Clustering via Fast Propagation of Cluster-wise Similarities}

\author{Dong~Huang,~\IEEEmembership{Member,~IEEE,}
        Chang-Dong~Wang,~\IEEEmembership{Member,~IEEE,}
        Hongxing Peng,\\
        Jianhuang~Lai,~\IEEEmembership{Senior Member,~IEEE}
        and~Chee-Keong Kwoh,~\IEEEmembership{Senior Member,~IEEE}
\thanks{This project was supported by NSFC (61602189, 61502543, 61573387 \& 61876193), National Key Research and Development Program of China (2016YFB1001003), Natural Science Foundation of Guangdong Province (2016A030310457 \& 2016A030306014), Singapore Ministry of Education Tier-1 Grant (RG21/15), and Singapore Ministry of Education Tier-2 Grant (MOE2014-T2-2-023).}
\thanks{D. Huang is with the College of Mathematics and Informatics, South China Agricultural University, Guangzhou, China, and also with the School of Computer Science and Engineering, Nanyang Technological University, Singapore. E-mail: huangdonghere@gmail.com.}
\thanks{C.-D. Wang is with the School of Data and Computer Science,
Sun Yat-sen University, Guangzhou, China, and also with Guangdong Province Key Laboratory of Computational Science, Guangzhou, China. E-mail: changdongwang@hotmail.com.}
\thanks{H. Peng is with the College of Mathematics and Informatics, South China Agricultural University, Guangzhou, China. E-mail: xyphx@scau.edu.cn.}
\thanks{J. Lai is with the School of Data and Computer Science,
Sun Yat-sen University, Guangzhou, China, and also with Guangdong Key Laboratory of Information Security Technology, Guangzhou, China, and also with Key Laboratory of Machine Intelligence and Advanced Computing, Ministry of Education, China. E-mail: stsljh@mail.sysu.edu.cn.}
\thanks{C.-K. Kwoh is with the School of Computer Science and Engineering, Nanyang Technological University, Singapore. E-mail: asckkwoh@ntu.edu.sg.}}

\markboth{IEEE Transactions on Systems, Man, and Cybernetics: Systems}%
{}

\maketitle

\begin{abstract}
Ensemble clustering has been a popular research topic in data mining and machine learning. Despite its significant progress in recent years, there are still two challenging issues in the current ensemble clustering research. First, most of the existing algorithms tend to investigate the ensemble information at the object-level, yet often lack the ability to explore the rich information at higher levels of granularity. Second, they mostly focus on the direct connections (e.g., direct intersection or pair-wise co-occurrence) in the multiple base clusterings, but generally neglect the multi-scale indirect relationship hidden in them. To address these two issues, this paper presents a novel ensemble clustering approach based on fast propagation of cluster-wise similarities via random walks. We first construct a cluster similarity graph with the base clusters treated as graph nodes and the cluster-wise Jaccard coefficient exploited to compute the initial edge weights. Upon the constructed graph, a transition probability matrix is defined, based on which the random walk process is conducted to propagate the graph structural information. Specifically, by investigating the propagating trajectories starting from different nodes, a new cluster-wise similarity matrix can be derived by considering the trajectory relationship. Then, the newly obtained cluster-wise similarity matrix is mapped from the cluster-level to the object-level to achieve an enhanced co-association (ECA) matrix, which is able to simultaneously capture the object-wise co-occurrence relationship as well as the multi-scale cluster-wise relationship in ensembles. Finally, two novel consensus functions are proposed to obtain the consensus clustering result. Extensive experiments on a variety of real-world datasets have demonstrated the effectiveness and efficiency of our approach.
\end{abstract}


\IEEEpeerreviewmaketitle

\section{Introduction}

\IEEEPARstart{D}{ata} clustering is an unsupervised learning technique that aims to partition a set of data objects (i.e., data points) into a certain number of homogeneous groups \cite{frey07_ap,das08_tsmca,meap13,svstream13,yang15,wang16_tkde,Chen18_tsmcs,Zhang18_tsmcs,He18_tsmcs,wu17_Euler}. It is a fundamental yet very challenging topic in the field of data mining and machine learning, and has been successfully applied in a wide variety of areas, such as image processing \cite{jm00_ncut,Huang16_neucom}, community discovery \cite{Wang14_tsmcs,neiwalk14_tkde}, recommender systems \cite{rafa13_tsmcs,symeon16_tsmcs,zhao18_tsmcs} and text mining \cite{rajp14_tsmcs}. In the past few decades, a large number of clustering algorithms have been developed by exploiting various techniques \cite{jain10_survey}. Different algorithms may lead to very different clustering performances for a specific dataset. Each clustering algorithm has its own advantages as well as weaknesses. However, there is no single algorithm that is suitable for all data distributions and applications. Given a clustering task, it is generally not easy to choose a proper clustering algorithm for it, especially without prior knowledge. Even if a specific algorithm is given, it may still be very difficult to decide the optimal parameters for the clustering task.

Unlike the conventional practice that typically uses a single algorithm to produce a single clustering result, ensemble clustering has recently emerged as a powerful tool whose purpose is to combine multiple different clustering results (generated by different algorithms or the same algorithm with different parameter settings) into a probably better and more robust consensus clustering \cite{Fred05_EAC}. Ensemble clustering has been gaining increasing attention, and many ensemble clustering algorithms have been proposed in recent years \cite{huang14_weac,Yu14_pr,wu15_TKDE,huang15_ecfg,Huang16_TKDE,huang17_tcyb,Yu16_tkde_incremental,Kang16_kbs,huang17_iconip,liu17_tkde,Yu17_tkde,yu17_tcyb}. Despite its significant progress, there are still two challenging issues in the current research. First, most of the existing ensemble clustering algorithms investigate the ensemble information at the object-level, and often fail to explore the higher-level information in the ensemble of multiple base clusterings. Second, they mostly focus on the direct relationship in ensembles, such as direct intersections and pair-wise co-occurrence, but generally neglect the multi-scale indirect connections in the base clusterings, which may exhibit a negative influence on the robustness of their consensus clustering performances.

\begin{figure}[!t]
\begin{center}
{\subfigure[]
{\includegraphics[width=0.3\linewidth]{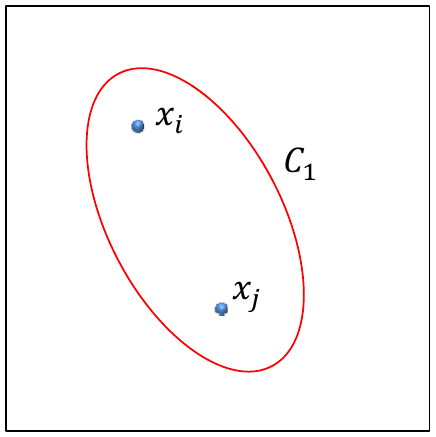}\label{fig:exampleEnsemble1}}}
{\subfigure[]
{\includegraphics[width=0.3\linewidth]{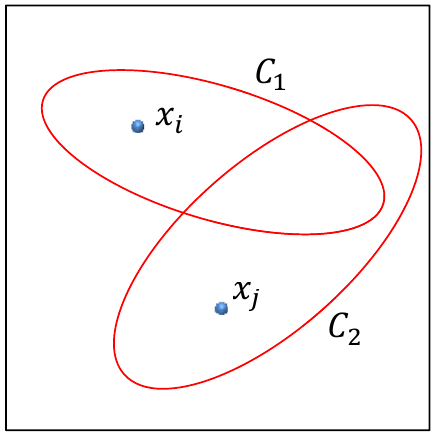}\label{fig:exampleEnsemble2}}}
{\subfigure[]
{\includegraphics[width=0.3\linewidth]{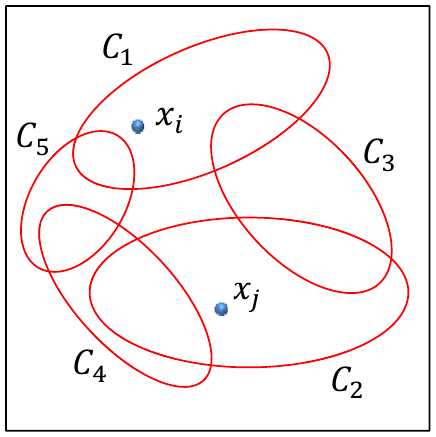}\label{fig:exampleEnsemble3}}}
\caption{The relationship between two objects $x_i$ and $x_j$ (a) if they appear in the same cluster, (b) if they appear in two different but intersected clusters, and (c) if they appear in two different clusters that are indirectly connected by some other clusters.}
\label{fig:exampleEnsemble}
\end{center}
\end{figure}

\begin{figure}[!t]
\begin{center}
{\subfigure[]
{\includegraphics[width=0.298\linewidth]{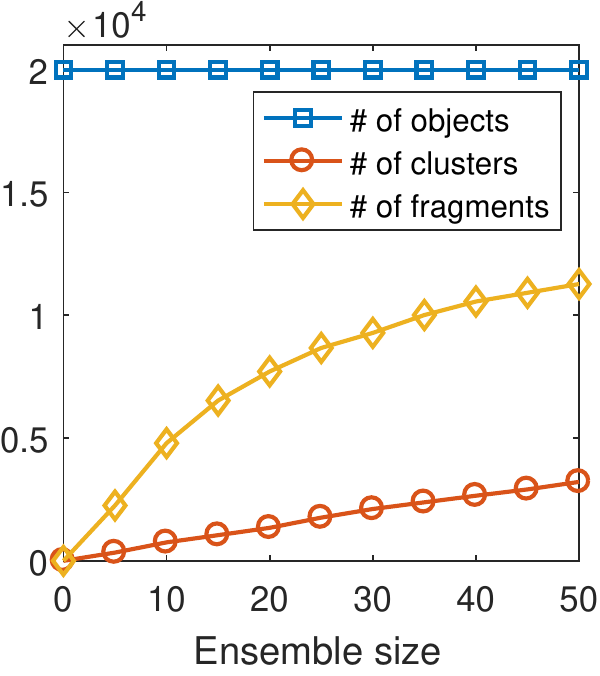}\label{fig:cmpMcSize1}}}
{\subfigure[]
{\includegraphics[width=0.336\linewidth]{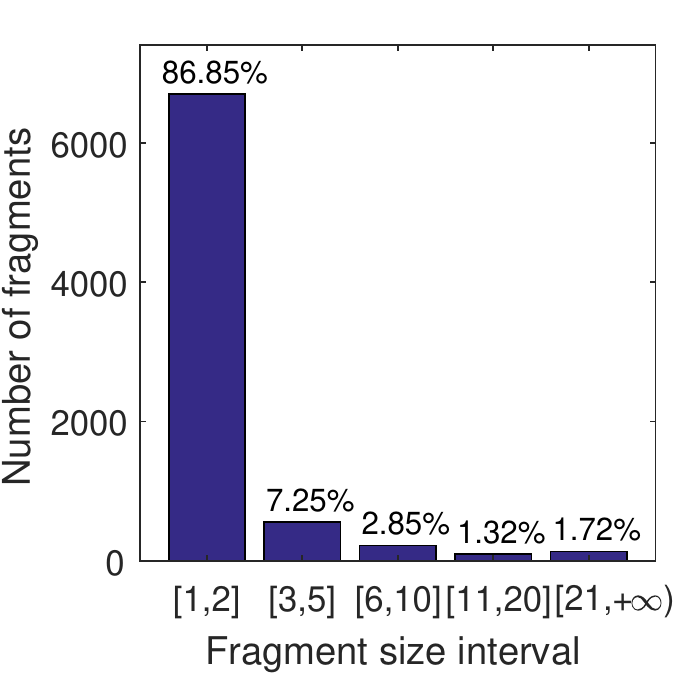}\label{fig:cmpMcSize2}}}
{\subfigure[]
{\includegraphics[width=0.336\linewidth]{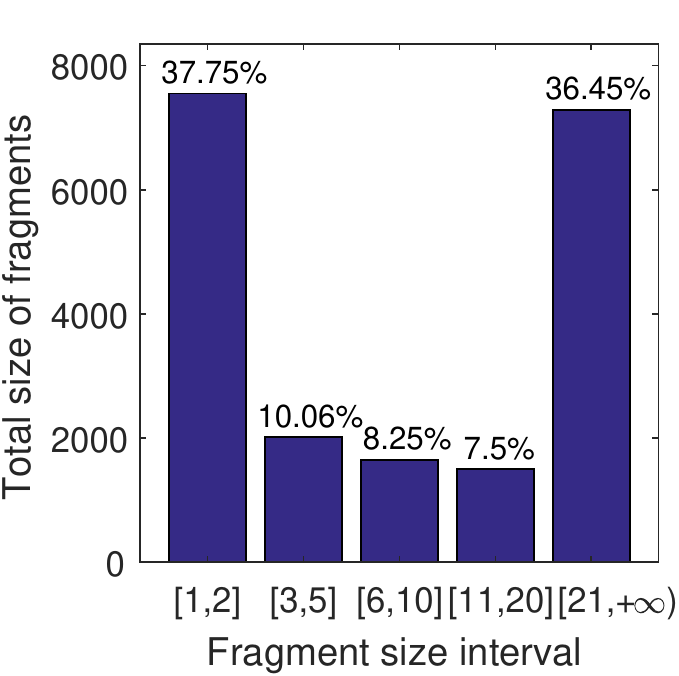}\label{fig:cmpMcSize3}}}
\caption{Statistics of the intersection fragments on the \emph{LR} datasets. (a) Numbers of data objects, clusters, and fragments as the ensemble size $M$ increases from 0 to 50. (b) Number of fragments in each size interval (with $M=20$). (c) Total size of fragments in each size interval (with $M=20$).}
\label{fig:cmpMcSize}
\end{center}
\end{figure}

In ensemble clustering, the direct co-occurrence relationship between objects is the most basic information. Fred and Jain \cite{Fred05_EAC} captured the co-occurrence relationship by presenting the concept of co-association matrix, which reflects how many times two objects occur in the \emph{same} cluster among the multiple base clusterings. The drawback of the conventional co-association matrix lies in that it only considers the direct co-occurrence relationship, yet lacks the ability to take into consideration the rich information of indirect connections in ensembles. As shown in Fig.~\ref{fig:exampleEnsemble}, if two objects occur in the same cluster in a base clustering, then we say these two objects are directly connected. If two objects are in two different clusters \emph{and} the two clusters are directly or indirectly related to each other, then we say these two objects are indirectly connected. The challenge here is two-fold: (i) how to improve the object-wise relationship by exploiting the higher-level (e.g., cluster-level) connections; (ii) how to explore the direct and indirect structural relationship in a unified model. To partially address this, Iam-On et al. \cite{iam_on11_linkbased} proposed the weighted connected triple (WCT) method to incorporate the common neighborhood information between clusters into the conventional co-association matrix, which exploits the direct neighborhood information between clusters but cannot utilize their indirect neighboring connections. Further, Iam-On et al. \cite{iamon08_icds} took advantage of the SimRank similarity (SRS) to investigate the indirect connections for refining the co-association matrix, which, however, suffers from its very high computational cost and is not feasible for large datasets. More recently, Huang et al. \cite{Huang16_TKDE} investigated the ensemble information by performing the random walk on a set of data fragments (also known as microclusters). Specifically, the set of data fragments are generated by intersecting the cluster boundaries of multiple base clusterings, and can be used as a set of basic operating units in the consensus process \cite{Huang16_TKDE}. Although using data fragments instead of original objects may provide better computational efficiency, the approach in \cite{Huang16_TKDE} still suffers from two limitations. In one aspect, when the ensemble size (i.e., the number of base clusterings) grows larger, the number of the generated fragments may increase dramatically (as shown in Fig.~\ref{fig:cmpMcSize1}), which eventually leads to a rapidly increasing computational burden. In another aspect, by intersecting the cluster boundaries of multiple base clusterings, the generated fragments may be associated with very \emph{imbalanced} sizes. As an example, we use twenty base clusterings on the \emph{Letter Recognition} (\emph{LR}) dataset to generate a set of data fragments. Different intersection fragments may have very different sizes, i.e., they may consist of very different numbers of data objects. The number of the fragments in each size interval is illustrated in Fig.~\ref{fig:cmpMcSize2}, while the total size of fragments in each size interval is shown in Fig.~\ref{fig:cmpMcSize3}. It can be observed that over 80 percent of the fragments have a very small size (of 1 or 2), whereas only 1.72 percent of the total fragments have a size greater than 20. However, the 1.72 percent of these large fragments surprisingly amounts to as large as 36.45 percent of the entire set of objects, which shows the heavy imbalance of the fragment sizes and places an unstable factor on the overall consensus process. Despite the efforts that these algorithms have made \cite{Huang16_TKDE,iam_on11_linkbased,iamon08_icds}, it remains an open problem how to effectively and efficiently investigate the higher-level ensemble information as well as incorporate multi-scale direct and indirect connections in ensembles for enhancing the consensus clustering performance.

To address the aforementioned challenging issues, in this paper, we propose a new ensemble clustering approach based on fast propagation of cluster-wise similarities via random walks. Different from the existing techniques that work at the object-level \cite{Fred05_EAC} or the fragment-level \cite{Huang16_TKDE}, in this paper, we explore the rich information of the ensembles at the base-cluster-level with multi-scale integration and cluster-object mapping. Specifically, a cluster similarity graph is first constructed by treating the base clusters as graph nodes and using the Jaccard coefficient to build the weighted edges. By defining a transition probability matrix, the random walk process is then performed to explore the multi-scale structural information in the cluster similarity graph. Thereafter, a new cluster-wise similarity matrix can be derived by utilizing the random walk trajectories starting from different nodes in the original graph. Further, an enhanced co-association (ECA) matrix is constructed by mapping the newly obtained cluster-wise similarity back to the object-level. Finally, by performing the partitioning process at the object-level and at the cluster-level, respectively, two novel consensus functions are therefore proposed, i.e., ensemble clustering by propagating cluster-wise similarities with hierarchical consensus function (ECPCS-HC) and ensemble clustering by propagating cluster-wise similarities with meta-cluster based consensus function (ECPCS-MC). Extensive experiments have been conducted on a variety of real-world datasets, which demonstrate the effectiveness and efficiency of our ensemble clustering approach when compared to the state-of-the-art approaches.

For clarity, the main contributions of this paper are summarized as follows:

\begin{itemize}
  \item A new cluster-wise similarity measure is derived, which captures the higher-level ensemble information and incorporates the multi-scale indirect connections by means of the random walk process starting from each cluster node.
  \item An enhanced co-association matrix is presented based on the cluster-object mapping, which simultaneously reflects the object-wise co-occurrence relationship as well as the cluster-wise structural information.
  \item Two novel consensus functions are devised, namely, ECPCS-HC and ECPCS-MC, which perform the partitioning process at the object-level and at the cluster-level, respectively, to obtain the final consensus clustering.
  \item Experiments on multiple datasets have shown the superiority of the proposed approach over the existing ensemble clustering approaches.
\end{itemize}

The remainder of the paper is organized as follows. Section~\ref{sec:related_work} reviews the related work on ensemble clustering. Section~\ref{sec:formulation} provides the formulation of the ensemble clustering problem. Section~\ref{sec:propagation_of_CSG} describes the construction of the cluster similarity graph and the random walk propagation for multi-scale integration. Section~\ref{sec:mapping_similarity} presents the enhanced co-association matrix by mapping cluster-wise similarities to object-wise similarities. Section~\ref{sec:consensus_function} proposes two novel consensus functions in our cluster-wise similarity propagation framework. Section~\ref{sec:experiments} reports the experimental results. Finally, Section~\ref{sec:conclusion} concludes this paper.

\section{Related Work}
\label{sec:related_work}

Ensemble clustering aims to combine a set of multiple base clusterings into a better and more robust consensus clustering result \cite{Fred05_EAC}. In the past decade, many ensemble clustering algorithms have been proposed \cite{huang15_ecfg,Huang16_TKDE,huang17_tcyb,huang17_iconip,liu17_tkde,Yu17_tkde,strehl02,fern04_bipartite,topchy05,li07,Li_WCC08,Mimaroglu11_pr,yi_icdm12,franek13_pr,Zhong15_pr,liu17_bioinformatics}, which can be classified into three main categories, namely, the pair-wise co-occurrence based algorithms \cite{Fred05_EAC,li07,yi_icdm12}, the graph partitioning based algorithms \cite{strehl02,fern04_bipartite,Mimaroglu11_pr}, and the median partition based algorithms \cite{huang15_ecfg,topchy05,Li_WCC08,franek13_pr}.

The pair-wise co-occurrence based algorithms \cite{Fred05_EAC,li07,yi_icdm12} typically build a co-association matrix by considering the frequency that two objects occur in the same cluster among the multiple base clusterings. By treating the co-association matrix as the similarity matrix, the hierarchical agglomerative clustering algorithms \cite{jain10_survey} can be used to obtain the consensus result. Fred and Jain \cite{Fred05_EAC} for the first time presented the concept of co-association matrix and designed the evidence accumulation clustering (EAC) method. Then, Li et al. \cite{li07} extended the EAC method by presenting a new hierarchical agglomerative clustering algorithm that takes the sizes of clusters into consideration via normalized edges. Yi et al. \cite{yi_icdm12} dealt with the uncertain entries in the co-association matrix by exploiting the matrix completion technique to improve the robustness of the consensus clustering.

The graph partitioning based algorithms \cite{strehl02,fern04_bipartite,Mimaroglu11_pr} formulate the clustering ensemble into a graph model and obtain the consensus clustering by segmenting the graph into a certain number of subsets. Strehl and Ghosh \cite{strehl02} treated each cluster in the set of base clusterings as a hyper-edge and proposed three graph partitioning based ensemble clustering algorithms. Fern and Brodley \cite{fern04_bipartite} built a bipartite graph by treating both clusters and objects as graph nodes, which is then partitioned via the METIS algorithm to obtain the consensus clustering. Mimaroglu and Erdil \cite{Mimaroglu11_pr} constructed a similarity graph between data objects and partitioned the graph by finding pivots and growing clusters.

The median partition based algorithms cast the ensemble clustering problem into an optimization problem which aims to find a median partition (or clustering) such that the similarity between this clustering and the set of base clusterings is maximized \cite{huang15_ecfg,topchy05,Li_WCC08,franek13_pr}. To deal with the median partition problem, which is NP-hard \cite{topchy05}, Topchy et al. \cite{topchy05} utilized the EM algorithm to find an approximate solution for it. Li et al. \cite{Li_WCC08} formulated the ensemble clustering problem into a nonnegative matrix factorization (NMF) problem and proposed the weighted consensus clustering (WCC) method. Franek et al. \cite{franek13_pr} cast the ensemble clustering problem into an Euclidean median problem and obtained an approximate solution via the Weiszfeld algorithm \cite{Weiszfeld09}. Huang et al. \cite{huang15_ecfg} formulated the ensemble clustering problem into a binary linear programming problem and solved it via the factor graph model \cite{Kschischang_FG_SPA:01}.

Despite the fact that significant progress has been made in the ensemble clustering research in recent years \cite{huang15_ecfg,Huang16_TKDE,huang17_tcyb,strehl02,fern04_bipartite,topchy05,li07,Li_WCC08,Mimaroglu11_pr,yi_icdm12,franek13_pr,ren13_icdm,yu15_tkde,yu15_tcbb,Ren17_kais}, there are still two challenging issues in most of the existing algorithms. First, they mostly investigate the ensemble information at the object-level, but often fail to go beyond the object-level to explore the information at higher levels of granularity in the ensemble. Second, many of them only consider the direct connections in ensembles and lack the ability to incorporate the multi-scale (indirect) connections for improving the consensus robustness. To (partially) address this, Iam-On et al. \cite{iam_on11_linkbased} proposed to refine the co-association matrix by considering the common neighborhood information between clusters, which in fact exploits the one-step indirect connections yet still neglects the multi-step (or multi-scale) indirect connections in ensembles. Further, Iam-On et al. \cite{iamon08_icds} exploited the SimRank Similarity (SRS) to incorporate the multi-scale neighborhood information in ensembles, which unfortunately suffers from its very high computational cost and is not feasible for large datasets. Huang et al. \cite{Huang16_TKDE} proposed to explore the structural information in ensembles by conducting random walks on the data fragments that are generated by intersecting the cluster boundaries of multiple base clusterings. However, in one aspect, the number of fragments would increase dramatically as the number of base clusterings grows larger, which may bring in a very heavy computational burden \cite{Huang16_TKDE}. In another aspect, the potentially imbalanced nature of the fragments (as shown in Figs.~\ref{fig:cmpMcSize2} and \ref{fig:cmpMcSize3}) also places an unstable factor on the robustness of the overall consensus clustering process. Moreover, while working at the fragment-level, the approach in \cite{Huang16_TKDE} still lacks the desired ability to investigate the multi-scale cluster-wise relationship in ensembles. Although considerable efforts have been made \cite{Huang16_TKDE,iam_on11_linkbased,iamon08_icds}, it remains a very challenging task how to simultaneously tackle the aforementioned two issues effectively and efficiently for the ensemble clustering problem.

\section{Problem Formulation}
\label{sec:formulation}

Ensemble clustering is the process of combining multiple base clusterings into a better consensus clustering result. Let $\mathcal{X}=\{x_1,\cdots,x_N\}$ denote a dataset with $N$ objects, where $x_i$ is the $i$-th object. Let $\Pi=\{\pi^1,\cdots,\pi^M\}$ denote a set of $M$ base clusterings for the dataset, where $\pi^m=\{C_1^m,\cdots,C_{n^m}^m\} $
is the $m$-th base clustering, $C_i^m$ is the $i$-th cluster in $\pi^m$, and $n^m$ is the number of clusters in $\pi^m$.

For clarity, the set of all clusters in the clustering ensemble $\Pi$ is denoted as $\mathcal{C} = \{C_1,\cdots,C_{N_c}\}$,
where $C_i$ is the $i$-th cluster and $N_c$ is the total number of clusters in $\Pi$. Obviously, it holds that $N_c=\sum_{m=1}^Mn^m$.

Formally, the objective of ensemble clustering is to integrate the information of the ensemble of multiple base clusterings in $\Pi$ to build a better clustering result $\pi^*$.

\section{Propagation of Cluster-wise Similarities}
\label{sec:propagation_of_CSG}

In ensemble clustering,  each base clustering consists of a certain number of base clusters. To capture the base cluster information, a commonly used strategy is to map the base cluster labels to the object-level \cite{Fred05_EAC} (or fragment-level \cite{Huang16_TKDE}) by building a co-association matrix, which reflects how many times two objects (or two fragments) are grouped in the same cluster among the multiple base clusterings. The straightforward mapping from the base cluster labels to the object-wise (or fragment-wise) co-association matrix implicitly assumes that different clusters are independent of each other, but fails to consider the potentially rich information hidden in the relationship between different clusters. In light of this, we aim to effectively and efficiently investigate the multi-scale direct and indirect relationship between base clusters in the ensemble, so as to achieve better and more robust consensus clustering results. Toward this end, two sub-problems here should first be solved, i.e., (i) how to define the initial similarity between clusters and (ii) how to incorporate the multi-scale information to construct more robust cluster-wise similarity.

Since a cluster is a set of data objects, the initial relationship between clusters can be investigated by the Jaccard coefficient \cite{Levandowsky1971}, which measures the similarity between two sets by considering their intersection size and union size. Formally, the Jaccard coefficient between two clusters (or sets), say, $C_i$ and $C_j$, is computed as \cite{Levandowsky1971}
\begin{equation}
\label{eq:eij}
Jaccard(C_i,C_j)=\frac{|C_i\bigcap C_j|}{|C_i\bigcup C_j|},
\end{equation}
where $\bigcap$ denotes the intersection of two sets, $\bigcup$ denotes the union of two sets, and $|\cdot|$ denotes the number of objects in a set. By adopting the Jaccard coefficient as the similarity measure between clusters, an initial cluster similarity graph is constructed for the ensemble with each cluster treated as a graph node. That is
\begin{equation}
\label{eq:cls_graph}
\mathcal{G}=\{\mathcal{V}, \mathcal{E}\},
\end{equation}
where $\mathcal{V}=\mathcal{C}$ is the node set and $\mathcal{E}$ is the edge set in the graph $\mathcal{G}$. The weight of an edge between two nodes $C_i,C_j\in\mathcal{V}$ is computed as
\begin{equation}
\label{eq:eij}
e_{ij}=Jaccard(C_i,C_j),
\end{equation}

With the initial similarity graph constructed, the next step is to incorporate the multi-scale information in the graph to enhance the cluster-wise similarity. In particular, the random walk process is performed on the graph, which is a dynamic process that transits from a node to one of its neighbors at each step with a certain probability \cite{neiwalk14_tkde,Huang16_TKDE,lovasz1993random,newman04,pon05_rw,lai_PRE10}. It is a crucial task in random walk to construct the transition probability matrix, which decides the probability of the random walker transiting from a node to another one. In this paper, the transition probability matrix $P=\{p_{ij}\}_{N\times N}$ on the graph is computed as follows:
\begin{align}
\label{eq:pij}
p_{ij} = \begin{cases}\frac{e_{ij}}{\sum_{C_k\neq C_i}e_{ik}},&\text{if~} i\neq j,\\
0, &\text{if~} i=j,
\end{cases}
\end{align}
where $p_{ij}$ is the probability that a random walker transits from node $C_i$ to node $C_j$ in one step, which is proportional to the edge weight between them. Based on the one-step transition probability matrix, we can obtain the multi-step transition probability matrix $P^{(t)}=\{p^{(t)}_{ij}\}_{N\times N}$ for the random walkers on the graph. That is
\begin{align}
P^{(t)}=\begin{cases}P,&\text{if~}t=1,\\
P^{(t-1)}\cdot P, &\text{if~}t>1.
\end{cases}
\end{align}
Note that the $(i,j)$-th entry in $P^{(t)}$, i.e., $p^{(t)}_{ij}$, denotes the probability of a random walker transiting from node $C_i$ to node $C_j$ in $t$ steps. We denote the $i$-th row in $P^{(t)}$ as $P^{(t)}_{i:}=\{p^{(t)}_{i1},p^{(t)}_{i2},\cdots,p^{(t)}_{iN}\}$, which represents the probability distribution of a random walker transiting from $C_i$ to all the other nodes in $t$ steps. As different step-lengths of random walkers can reflect the graph structure information at different scales \cite{Huang16_TKDE,lai_PRE10}, to capture the multi-scale information in the graph $\mathcal{G}$, the random walk trajectories at different steps are exploited here to refine the cluster-wise similarity.

Formally, for the random walker starting from a node $C_i$, its random walk trajectory from step 1 to step $t$ is denoted as
$P^{(1:t)}_{i:}=\{P^{(1)}_{i:},P^{(2)}_{i:},\cdots,P^{(t)}_{i:}\}$.
Obviously, the $t$-step random walk trajectory (i.e., $P^{(1:t)}_{i:}$), starting from node $C_i$ and having a step-length $t$, is an $N\cdot t$-tuple, which captures the multi-scale (or multi-step) structural information in the neighborhood of $C_i$. With the random walk trajectory of each node obtained, a new similarity measure can thereby be derived for every two nodes by considering the similarity of their random walk trajectories. Specifically, the new similarity matrix between all of the clusters in $\Pi$ is represented as
\begin{equation}
Z=\{z_{ij}\}_{N_c\times N_c},
\end{equation}
where
\begin{equation}
\label{eq:PTS}
z_{ij}=Sim(P^{(1:t)}_{i:},P^{(1:t)}_{j:}).
\end{equation}
denotes the new similarity between two clusters $C_i$ and $C_j$. Note that $Sim(\cdot,\cdot)$ can be any similarity measure between two vectors. In our paper, the cosine similarity \cite{tan2005introduction} is adopted. Thus, the new similarity measure between $C_i$ and $C_j$ can be computed as
\begin{align}
\label{eq:PTS_cos}
z_{ij}=\frac{<P^{(1:t)}_{i:},P^{(1:t)}_{j:}>}{\sqrt{<P^{(1:t)}_{i:},P^{(1:t)}_{i:}>\cdot <P^{(1:t)}_{j:},P^{(1:t)}_{j:}>}},
\end{align}
where $<\cdot,\cdot>$ outputs the dot product of two vectors. Since the entries in the transition probability matrix are always non-negative, it holds that $z_{ij}\in [0,1]$ for any two clusters $C_i$ and $C_j$ in $\Pi$.

\section{Enhanced Co-association Matrix Based on Similarity Mapping}
\label{sec:mapping_similarity}

Having obtained the new cluster-wise similarity matrix $Z$, we proceed to map the new similarity matrix from the cluster-level to the object-level, and describe the enhanced co-association representation in this section.

The conventional co-association matrix \cite{Fred05_EAC} is a widely used data structure to capture the object-wise similarity in the ensemble clustering problem. Given the clustering ensemble $\Pi$, the (direct) pair-wise relationship in the $m$-th base clustering (i.e., $\pi^m$) can be represented by a connectivity matrix, which is computed as follows:
\begin{align}
A^m &=\{a^m_{ij}\}_{N\times N},\\
a^m_{ij} &= \begin{cases} 1, &\text{if~}Cls^m(x_i)=Cls^m(x_j),\\
0, &\text{otherwise,}
\end{cases}
\end{align}
where $Cls^m(x_i)$ denotes the cluster in $\pi^m$ that contains the object $x_i$. Obviously, if $C_j\in\pi^m$ and $x_i\in C_j$, then $Cls^m(x_i)=C_j$. Then, the conventional co-association matrix $A=\{a_{ij}\}_{N\times N}$ for the entire ensemble is computed as follows:
\begin{align}
A = \frac{1}{M}\sum_{m=1}^{M}A^m.
\end{align}
The conventional co-association matrix reflects the number of times that two objects appear in the same cluster among the multiple base clusterings. Although it is able to exploit the (direct) cluster-level information by investigating the object-wise co-occurrence relationship, it inherently treats each cluster as an independent entity and, however, neglects the potential relationship between \emph{different} clusters, which may provide rich information for further refining the object-wise connections. In light of this, with the multi-scale cluster-wise relationship explored by random walks in Section~\ref{sec:propagation_of_CSG}, the key problem in this section is how to map the multi-scale cluster-wise relationship back to the object-level.

In particular, we present an enhanced co-association (ECA) matrix to simultaneously capture the object-wise co-occurrence relationship and the multi-scale cluster-wise similarity. Before the construction of the ECA matrix for the entire ensemble, we first take advantage of the newly designed cluster-wise similarity matrix $Z$ to build the enhanced connectivity matrix for a single base clusterings, say, $\pi^m$. That is
\begin{align}
B^m &=\{b^m_{ij}\}_{N\times N},\\
b^m_{ij} &= \begin{cases} 1, &\text{if~}Cls^m(x_i)=Cls^m(x_j),\\
z_{uv}, &\text{if~}Cls^m(x_i)\neq Cls^m(x_j),\\
\end{cases}
\end{align}
with
\begin{align}
Cls^m(x_i)&=C^m_u, Cls^m(x_j)=C^m_v.
\end{align}
Note that the $(i,j)$-th entry in $B^m$ and the $(i,j)$-th entry in $A^m$ will be the same \emph{only when} $x_i$ and $x_j$ occur in the same cluster in $\pi^m$. The difference between $B^m$ and $A^m$ arises when two objects belongs to different clusters in a base clustering, in which situation the conventional connectivity matrix $A^m$ lacks the ability to go beyond the direct co-occurrence relationship to exploit further cluster-wise connections. Different from the convectional connectivity matrix, when two objects belong to two different clusters in a base clustering, the enhanced connectivity matrix $B^m$ is still able to capture their indirect relationship by investigating the correlation between the two clusters that these two objects respectively belong to.

With the enhanced connectivity matrix for each base clustering constructed, the ECA matrix, denoted as $B=\{b_{ij}\}_{N\times N}$, for the entire ensemble $\Pi$ can be computed as follows:
\begin{align}
B = \frac{1}{M}\sum_{m=1}^{M}B^m.
\end{align}
With $z_{ij}\in [0,1]$, it is obvious that all entries in the ECA matrix are in the range of $[0,1]$. By the construction of the ECA matrix, the cluster-wise similarity in $Z$ is mapped from the cluster-level to the object-level. It is noteworthy that the ECA matrix can be utilized in any co-association matrix based consensus functions. In particular, two new consensus functions will be designed in the next section.

\section{Two Types of Consensus Functions}
\label{sec:consensus_function}

In this section, we propose two consensus functions to obtain the final consensus clustering in the proposed ensemble clustering by propagating cluster-wise similarities (ECPCS) framework. The first consensus function is described in Section~\ref{sec:ECPCS_HC}, which is based on hierarchical clustering and performs the partitioning process at the object-level, while the second consensus function is described in Section~\ref{sec:ECPCS_MC}, which is based on meta-clustering and performs the partitioning process at the cluster-level.

\subsection{ECPCS-HC}
\label{sec:ECPCS_HC}

In this section, we describe our first consensus function termed ECPCS-HC, short for ECPCS with hierarchical consensus function. By treating the ECA matrix as the new object-wise similarity matrix, the hierarchical agglomerative clustering can be performed to obtain the consensus clustering in an iterative region merging fashion. The original objects are viewed as the set of initial regions, that is
\begin{align}
\mathcal{R}^{(0)}=\{R^{(0)}_1, \cdots, R^{(0)}_N\},
\end{align}
where $R^{(0)}_i=\{x_i\}$ denotes the $i$-th initial region that contains exactly one object $x_i$. The similarity matrix for the set of initial regions is defined as
\begin{align}
S^{(0)}&=\{s^{(0)}_{ij}\}_{N\times N},\\
s^{(0)}_{ij}&=b_{ij}.
\end{align}
With the initial region set and its similarity matrix obtained, the region merging process is then performed iteratively. In each iteration, the two regions with the highest similarity are merged into a new and larger region, which will be followed by the update of the region set and the corresponding similarity matrix. Specifically, the updated region set after the $q$-th iteration is denoted as
\begin{align}
\mathcal{R}^{(q)}=\{R^{(q)}_1, \cdots, R^{(q)}_{|\mathcal{R}^{(q)}|}\},
\end{align}
where $R^{(q)}_i$ is the $i$-th region and $|\mathcal{R}^{(q)}|$ is the number of regions in $\mathcal{R}^{(q)}$.

The similarity matrix after the $q$-th iteration is updated according to the average-link. That is
\begin{align}
S^{(q)}&=\{s^{(q)}_{ij}\}_{|\mathcal{R}^{(q)}|\times |\mathcal{R}^{(q)}|},\\
s^{(q)}_{ij}&=\frac{1}{|R^{(q)}_i|\cdot |R^{(q)}_j|}\sum_{x_u\in R^{(q)}_i,x_v\in R^{(q)}_j} b_{uv},
\end{align}
where $|R^{(q)}_i|$ denotes the number of objects in $R^{(q)}_i$.

Note that in each iteration the number of regions decreases by one, i.e., $|\mathcal{R}^{(q+1)}|=|\mathcal{R}^{(q)}|-1$. Since the number of the initial regions is $N$, it is obvious that all objects will be merged into a root region after totally $N-1$ iterations. As the result of the region merging process, a dendrogram (i.e., a hierarchical clustering tree) will be iteratively constructed. Each level of the dendrogram corresponds to a clustering result with a certain number of clusters. By choosing a level in the dendrogram,  the final consensus clustering can thereby be obtained.

\subsection{ECPCS-MC}
\label{sec:ECPCS_MC}

In this section, we describe our second consensus function termed ECPCS-MC, short for ECPCS with meta-cluster based consensus function. Different from ECPCS-HC, the ECPCS-MC method performs the partitioning process at the cluster-level, which takes advantage of the enhanced cluster-wise similarity matrix $Z$ and groups all the clusters in the ensemble into several subsets. Each subset of clusters is referred to as a meta-cluster. Then, each data object is assigned to one of the meta-clusters by majority voting to construct the final consensus clustering.

Specifically, by treating the clusters in the ensemble as graph nodes and using the cluster-wise similarity matrix $Z$ to define the edge weights between them, a new cluster similarity graph can be constructed. That is
\begin{equation}
\label{eq:cls_graph}
\tilde{\mathcal{G}}=\{\mathcal{V}, \tilde{\mathcal{E}}\},
\end{equation}
where $\mathcal{V}=\mathcal{C}$ is the node set and $\tilde{\mathcal{E}}$ is the edge set. The edge weights in the graph $\tilde{\mathcal{G}}$ are decided by the enhanced cluster-wise similarity matrix $B$. Given two clusters $C_i$ and $C_j$, the weight between them is defined as
\begin{equation}
\tilde{e}_{ij}=b_{ij}.
\end{equation}

Then, the normalized cut (Ncut) algorithm \cite{jm00_ncut} can be used to partition the new graph into a certain number of meta-clusters, that is
\begin{align}
\mathcal{MC}=\{MC_1,MC_2,\cdots,MC_k\},
\end{align}
where $MC_i$ is the $i$-the meta-cluster and $k$ is the number of meta-clusters.

Note that a meta-cluster consists of a certain number of clusters. Given an object $x_i$ and a meta-cluster $MC_j$, the object $x_i$ may appear in \emph{zero or more} clusters inside $MC_j$. Specifically, the voting score of $x_i$ w.r.t. the meta-cluster $MC_j$ can be defined as the proportion of the clusters in $MC_j$ that contain $x_i$. That is
\begin{align}
Score(x_i,MC_j)&=\frac{1}{|MC_j|}\sum_{C_l\in MC_j}\textbf{1}(x_i\in C_l),\\
\textbf{1}(statement)&=\begin{cases}1,&\text{if~$statement$~is~true},\\
0,&\text{otherwise}.\nonumber
\end{cases}
\end{align}
where $|MC_j|$ denotes the number of clusters in $MC_j$.

Then, by majority voting, each object is assigned to the meta-cluster in which it appears most frequently (i.e., with the highest voting score). That is
\begin{align}
MetaCls(x_i)={\arg\max}_{MC_j\in\mathcal{MC}}Score(x_i,MC_j).
\end{align}

If an object obtains the same highest voting score from two or more different meta-clusters (which in practice rarely happens), then the object will be randomly assigned to one of the winning meta-clusters. By assigning each object to a meta-cluster via majority voting and treating the objects in the same meta-cluster as a consensus cluster, the final consensus clustering result can therefore be obtained.

\section{Experiments}
\label{sec:experiments}

In this section, we conduct experiments on a variety of benchmark datasets to evaluate the performance of the proposed ECPCS-HC and ECPCS-MC algorithms against several state-of-the-art ensemble clustering algorithms.

\subsection{Datasets and Evaluation Measures}

In our experiments, ten benchmark datasets are used, i.e., \emph{Breast Cancer} (\emph{BC}), \emph{Cardiotocography} (\emph{CTG}), \emph{Ecoli}, \emph{Gisette}, \emph{Letter Recognition} (\emph{LR}), \emph{Landsat Satellite} (\emph{LS}), \emph{MNIST}, \emph{Pen Digits} (\emph{PD}), \emph{Wine}, and \emph{Yeast}. The \emph{MNIST} dataset is from \cite{lecun98}, while the other nine datasets are from the UCI machine learning repository \cite{Bache+Lichman:2013}. The detailed information of the benchmark datasets is given in Table~\ref{table:dataset}.

To quantitatively evaluate the clustering results, two widely used evaluation measures are adopted, namely, normalized mutual information (NMI) \cite{strehl02} and adjusted Rand index (ARI) \cite{vinh2010_ARI}. Note that large values of NMI and ARI indicate better clustering results.

The NMI evaluates the similarity between two clusterings from an information theory perspective \cite{strehl02}. Let $\pi'$ be a test clustering and $\pi^G$ be the ground-truth clustering. The NMI between $\pi'$ and $\pi^G$ is computed as follows \cite{strehl02}:
\begin{equation}
\label{eq:nmi}
NMI(\pi', \pi^G)=\frac{\sum_{i=1}^{n'}\sum_{j=1}^{n^G}n_{ij}\log\frac{n_{ij}n}{n_i'n_j^G}}{\sqrt{\sum_{i=1}^{n'}n_i'\log\frac{n_i'}{n}\sum_{j=1}^{n^G}n_j^G\log\frac{n_j^G}{n}}},
\end{equation}
where $n'$ is the cluster number in $\pi'$, $n^G$ is the cluster number in $\pi^G$, $n_i'$ is the number of objects in the cluster $i$ of $\pi'$, $n_j^G$ is the number of objects in the cluster $j$ of $\pi^G$, and $n_{ij}$ is the size of the intersection of the cluster $i$ of $\pi'$ and the cluster $j$ of $\pi^G$.

The ARI is an evaluation measure that takes into consideration the number of object-pairs upon which two clusterings agree (or disagree) \cite{vinh2010_ARI}. Formally, the ARI between two clusterings $\pi'$ and $\pi^G$ is computed as follows \cite{vinh2010_ARI}:
\begin{align}
&ARI(\pi', \pi^G)=\nonumber\\
& \frac{2(N_{00}N_{11}-N_{01}N_{10})}{(N_{00}+N_{01})(N_{01}+N_{11})+(N_{00}+N_{10})(N_{10}+N_{11})},
\end{align}
where $N_{11}$ is the number of object-pairs that belong to the same cluster in both $\pi'$ and $\pi^G$, $N_{00}$ is the number of object-pairs that belong to different clusters in both $\pi'$ and $\pi^G$, $N_{10}$ is the number of object-pairs that belong to the same cluster in  $\pi'$ while belonging to different clusters in $\pi^G$, and $N_{01}$ is the number of object-pairs that belong to different clusters in $\pi'$ while belonging to the same cluster in $\pi^G$.

\begin{table}[!t]\footnotesize
\centering
\caption{Description of the benchmark datasets.}
\label{table:dataset}
\begin{tabular}{m{1.99cm}<{\centering}|m{1.371cm}<{\centering}m{1.371cm}<{\centering}m{1.371cm}<{\centering}}
\toprule
Dataset &\#Object &\#Class &Dimension\\
\midrule
\emph{BC}  &683   &2   &9\\
\emph{CTG}  &2,126   &10   &21\\
\emph{Ecoli}  &336   &8   &7\\
\emph{Gisette}  &7,000   &2   &5,000\\
\emph{LR}  &20,000   &26   &16\\
\emph{LS}  &$6,435$   &6   &36\\
\emph{MNIST}  &$5,000$   &10   &784\\
\emph{PD}  &$10,992$   &10   &16\\
\emph{Wine}  &178   &3   &13\\
\emph{Yeast}  &1,484   &10   &8\\
\bottomrule
\end{tabular}
\end{table}

\begin{table*}[!t]
\centering
\caption{Average NMI($\%$) scores (over 20 runs) by different ensemble clustering methods. The best three scores in each comparison are highlighted in \textbf{bold}, while the best one in \textbf{[bold and brackets]}.}
\label{table:compare_ensembles_nmi}
\begin{tabular}{|m{1.16cm}<{\centering}|m{0.78cm}<{\centering}|m{1.09cm}<{\centering}m{1.15cm}<{\centering}m{1.09cm}<{\centering}m{1.09cm}<{\centering}m{1.09cm}<{\centering}m{1.15cm}<{\centering}m{1.09cm}<{\centering}m{1.25cm}<{\centering}|m{1.39065cm}<{\centering}m{1.344cm}<{\centering}|}
\hline
\multicolumn{1}{|c}{\emph{Dataset}} &\multicolumn{1}{c|}{} &EAC &MCLA &SRS &WCT &KCC &PTGP &ECC &SEC &ECPCS-MC &ECPCS-HC\\
\hline
\multirow{2}{*}{\emph{BC}}	&True-$k$	&73.00$_{\pm6.57}$	&77.63$_{\pm2.39}$	&72.46$_{\pm5.32}$	&76.32$_{\pm5.43}$	&76.18$_{\pm10.22}$	&76.09$_{\pm4.14}$	&\textbf{79.07}$_{\pm1.86}$	&45.26$_{\pm26.26}$	&\textbf{77.89}$_{\pm1.47}$	&[\textbf{79.46}$_{\pm3.47}$]\\
\cline{2-12}
	&Best-$k$	&73.34$_{\pm5.64}$	&77.63$_{\pm2.39}$	&72.56$_{\pm5.01}$	&76.33$_{\pm5.42}$	&76.59$_{\pm8.20}$	&76.09$_{\pm4.14}$	&\textbf{79.07}$_{\pm1.86}$	&54.58$_{\pm16.93}$	&\textbf{77.89}$_{\pm1.47}$	&[\textbf{79.46}$_{\pm3.47}$]\\
\hline
\multirow{2}{*}{\emph{CTG}}	&True-$k$	&\textbf{26.16}$_{\pm0.85}$	&24.71$_{\pm0.92}$	&25.85$_{\pm0.77}$	&26.15$_{\pm1.00}$	&23.28$_{\pm1.73}$	&25.15$_{\pm1.11}$	&23.58$_{\pm1.27}$	&24.41$_{\pm1.80}$	&[\textbf{26.87}$_{\pm0.91}$]	&\textbf{26.42}$_{\pm1.42}$\\
\cline{2-12}
	&Best-$k$	&26.47$_{\pm0.80}$	&25.63$_{\pm0.66}$	&26.27$_{\pm0.81}$	&\textbf{26.66}$_{\pm0.85}$	&25.07$_{\pm0.83}$	&26.38$_{\pm0.81}$	&24.78$_{\pm0.65}$	&25.44$_{\pm0.85}$	&[\textbf{27.60}$_{\pm0.76}$]	&\textbf{26.99}$_{\pm0.87}$\\
\hline
\multirow{2}{*}{\emph{Ecoli}}	&True-$k$	&58.33$_{\pm2.91}$	&49.17$_{\pm2.92}$	&56.79$_{\pm2.42}$	&\textbf{62.20}$_{\pm3.80}$	&49.64$_{\pm2.84}$	&50.28$_{\pm2.23}$	&50.61$_{\pm1.97}$	&51.76$_{\pm3.81}$	&\textbf{59.54}$_{\pm1.75}$	&[\textbf{70.48}$_{\pm2.51}$]\\
\cline{2-12}
	&Best-$k$	&68.02$_{\pm3.36}$	&52.68$_{\pm1.58}$	&67.40$_{\pm2.79}$	&\textbf{70.98}$_{\pm1.86}$	&54.68$_{\pm3.19}$	&60.63$_{\pm2.99}$	&57.63$_{\pm2.30}$	&55.01$_{\pm3.73}$	&\textbf{69.93}$_{\pm2.01}$	&[\textbf{71.45}$_{\pm0.96}$]\\
\hline
\multirow{2}{*}{\emph{Gisette}}	&True-$k$	&27.02$_{\pm13.60}$	&\textbf{41.69}$_{\pm12.52}$	&35.09$_{\pm9.52}$	&37.79$_{\pm8.35}$	&17.26$_{\pm12.74}$	&[\textbf{47.13}$_{\pm1.94}$]	&29.15$_{\pm10.08}$	&12.10$_{\pm7.97}$	&\textbf{47.01}$_{\pm2.23}$	&40.42$_{\pm8.25}$\\
\cline{2-12}
	&Best-$k$	&31.18$_{\pm8.78}$	&\textbf{43.13}$_{\pm8.37}$	&35.77$_{\pm8.39}$	&38.37$_{\pm7.17}$	&23.13$_{\pm7.57}$	&[\textbf{47.13}$_{\pm1.94}$]	&30.41$_{\pm7.64}$	&17.83$_{\pm5.70}$	&\textbf{47.01}$_{\pm2.23}$	&41.00$_{\pm7.05}$\\
\hline
\multirow{2}{*}{\emph{LR}}	&True-$k$	&38.30$_{\pm0.90}$	&38.60$_{\pm1.17}$	&38.40$_{\pm1.10}$	&38.48$_{\pm1.09}$	&34.87$_{\pm0.95}$	&\textbf{39.16}$_{\pm1.17}$	&35.72$_{\pm0.83}$	&33.13$_{\pm1.44}$	&[\textbf{39.30}$_{\pm0.74}$]	&\textbf{38.73}$_{\pm1.44}$\\
\cline{2-12}
	&Best-$k$	&41.64$_{\pm0.54}$	&40.53$_{\pm0.62}$	&42.14$_{\pm0.62}$	&\textbf{42.49}$_{\pm0.58}$	&38.78$_{\pm0.66}$	&41.85$_{\pm0.60}$	&39.22$_{\pm0.69}$	&38.88$_{\pm0.76}$	&[\textbf{42.85}$_{\pm0.55}$]	&[\textbf{42.85}$_{\pm0.84}$]\\
\hline
\multirow{2}{*}{\emph{LS}}	&True-$k$	&60.86$_{\pm3.73}$	&53.58$_{\pm3.16}$	&62.00$_{\pm3.80}$	&62.13$_{\pm2.59}$	&48.46$_{\pm3.67}$	&\textbf{62.45}$_{\pm1.33}$	&52.39$_{\pm4.21}$	&43.57$_{\pm6.97}$	&[\textbf{63.90}$_{\pm2.36}$]	&\textbf{63.18}$_{\pm2.55}$\\
\cline{2-12}
	&Best-$k$	&62.17$_{\pm2.17}$	&54.50$_{\pm2.30}$	&62.96$_{\pm1.59}$	&\textbf{63.79}$_{\pm1.61}$	&51.35$_{\pm2.41}$	&63.09$_{\pm1.18}$	&53.55$_{\pm3.33}$	&49.75$_{\pm3.58}$	&[\textbf{65.02}$_{\pm1.86}$]	&\textbf{64.86}$_{\pm1.34}$\\
\hline
\multirow{2}{*}{\emph{MNIST}}	&True-$k$	&61.94$_{\pm1.81}$	&58.26$_{\pm3.53}$	&\textbf{62.63}$_{\pm1.82}$	&62.44$_{\pm1.73}$	&50.90$_{\pm2.77}$	&\textbf{63.59}$_{\pm2.51}$	&50.02$_{\pm2.68}$	&45.74$_{\pm4.32}$	&[\textbf{63.81}$_{\pm2.15}$]	&60.26$_{\pm1.62}$\\
\cline{2-12}
	&Best-$k$	&62.47$_{\pm1.84}$	&58.60$_{\pm3.11}$	&62.99$_{\pm1.84}$	&63.73$_{\pm1.71}$	&54.13$_{\pm1.90}$	&\textbf{64.84}$_{\pm1.94}$	&53.54$_{\pm1.33}$	&55.12$_{\pm1.89}$	&\textbf{64.40}$_{\pm1.81}$	&[\textbf{65.00}$_{\pm1.36}$]\\
\hline
\multirow{2}{*}{\emph{PD}}	&True-$k$	&73.63$_{\pm2.16}$	&70.20$_{\pm3.37}$	&74.57$_{\pm2.67}$	&74.61$_{\pm3.13}$	&60.77$_{\pm3.74}$	&\textbf{74.80}$_{\pm3.38}$	&62.36$_{\pm2.67}$	&51.75$_{\pm7.58}$	&[\textbf{76.41}$_{\pm2.28}$]	&\textbf{74.91}$_{\pm3.24}$\\
\cline{2-12}
	&Best-$k$	&76.87$_{\pm1.24}$	&71.01$_{\pm2.75}$	&77.75$_{\pm1.38}$	&78.51$_{\pm1.72}$	&67.63$_{\pm1.94}$	&\textbf{79.11}$_{\pm1.54}$	&67.55$_{\pm1.50}$	&67.44$_{\pm2.11}$	&[\textbf{79.79}$_{\pm1.24}$]	&\textbf{78.84}$_{\pm1.65}$\\
\hline
\multirow{2}{*}{\emph{Wine}}	&True-$k$	&86.34$_{\pm2.60}$	&82.25$_{\pm3.16}$	&\textbf{88.05}$_{\pm2.88}$	&87.33$_{\pm3.11}$	&86.01$_{\pm3.69}$	&86.85$_{\pm2.51}$	&83.29$_{\pm7.10}$	&86.10$_{\pm4.13}$	&\textbf{87.85}$_{\pm2.36}$	&[\textbf{88.82}$_{\pm2.82}$]\\
\cline{2-12}
	&Best-$k$	&86.34$_{\pm2.60}$	&82.25$_{\pm3.16}$	&\textbf{88.05}$_{\pm2.88}$	&87.33$_{\pm3.11}$	&86.06$_{\pm3.41}$	&86.85$_{\pm2.51}$	&83.68$_{\pm6.19}$	&86.13$_{\pm4.01}$	&\textbf{87.85}$_{\pm2.36}$	&[\textbf{88.84}$_{\pm2.79}$]\\
\hline
\multirow{2}{*}{\emph{Yeast}}	&True-$k$	&26.21$_{\pm1.33}$	&22.12$_{\pm1.15}$	&26.03$_{\pm1.04}$	&\textbf{28.36}$_{\pm1.39}$	&21.54$_{\pm2.59}$	&23.42$_{\pm1.21}$	&19.53$_{\pm0.72}$	&22.02$_{\pm2.01}$	&\textbf{27.71}$_{\pm1.09}$	&[\textbf{29.62}$_{\pm1.21}$]\\
\cline{2-12}
	&Best-$k$	&28.44$_{\pm1.64}$	&23.15$_{\pm1.01}$	&28.05$_{\pm1.66}$	&\textbf{29.83}$_{\pm1.09}$	&23.37$_{\pm0.98}$	&27.76$_{\pm1.36}$	&24.30$_{\pm0.91}$	&23.49$_{\pm1.01}$	&\textbf{29.30}$_{\pm0.87}$	&[\textbf{30.05}$_{\pm0.97}$]\\
\hline
\hline
\multirow{2}{*}{Avg. score}	&True-$k$	&53.18	&51.82	&54.19	&55.58	&46.89	&54.89	&48.57	&41.58	&57.03	&57.23\\
\cline{2-12}
	&Best-$k$	&55.69	&52.91	&56.39	&57.80	&50.08	&57.37	&51.37	&47.37	&59.16	&58.93\\
\hline
\hline
\multirow{2}{*}{Avg. rank}	&True-$k$	&5.70	&6.60	&5.10	&3.90	&8.50	&4.30	&7.90	&8.80	&1.90	&2.30\\
\cline{2-12}
	&Best-$k$	&5.70	&7.20	&5.20	&3.60	&8.50	&4.30	&7.90	&8.70	&2.10	&1.70\\
\hline
\end{tabular}
\end{table*}

\begin{figure*}[!t]
\begin{center}
{\subfigure[]
{\includegraphics[width=0.895\columnwidth]{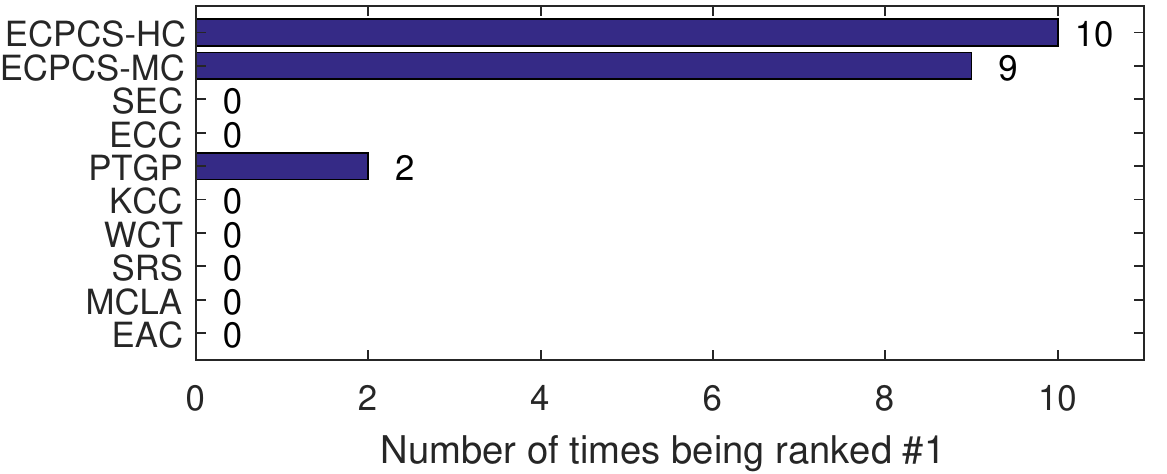}\label{fig:rankTop1Top3_nmi_1}}}
{\subfigure[]
{\includegraphics[width=0.895\columnwidth]{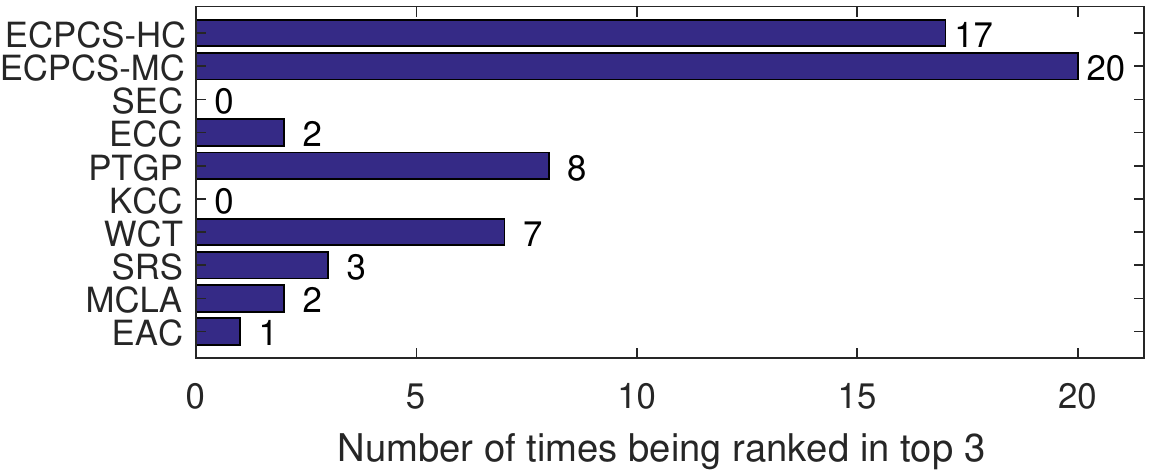}\label{fig:rankTop1Top3_nmi_2}}}
\caption{The number of times that each method is ranked (a) in the first position and (b) in the top three w.r.t. Table~\ref{table:compare_ensembles_nmi}.}
\label{fig:rankTop1Top3_nmi}
\end{center}
\end{figure*}

\subsection{Baseline Methods and Experimental Settings}

Our proposed ECPCS-HC and ECPCS-MC methods will be compared against eight baseline ensemble clustering methods, which are listed as follows:

\begin{enumerate}
  \item \textbf{EAC} \cite{Fred05_EAC}: evidence accumulation clustering.
  \item \textbf{MCLA} \cite{strehl02}: meta-clustering algorithm.
  \item \textbf{SRS} \cite{iamon08_icds}: SimRank similarity based method
  \item \textbf{WCT} \cite{iam_on11_linkbased}: weighted connected triple method.
  \item \textbf{KCC} \cite{wu15_TKDE}: $k$-means based consensus clustering.
  \item \textbf{PTGP} \cite{Huang16_TKDE}: probability trajectory based graph partitioning.
  \item \textbf{ECC} \cite{liu17_bioinformatics}: entropy based consensus clustering.
  \item \textbf{SEC} \cite{liu17_tkde}: spectral ensemble clustering.
\end{enumerate}

The parameters in the baseline methods are set as suggested by their corresponding papers \cite{Fred05_EAC,wu15_TKDE,Huang16_TKDE,liu17_tkde,iam_on11_linkbased,iamon08_icds,strehl02,liu17_bioinformatics}. The step-length parameter $t$ in the proposed methods is set to 20 for the experiments on all datasets, whose sensitivity will be further evacuated in Section~\ref{sec:sensitivity_t}.

To provide a fair comparison, we run each of the test methods twenty times on each dataset, and report their average NMI and ARI scores over multiple runs. At each run, an ensemble of $M=20$ base clusterings is constructed by the $k$-means clustering with initial cluster centers randomly initialized and the number of clusters in each base clustering randomly selected in the range of [K, min($\sqrt{N}$,100)], where $K$ is the number of classes and $N$ is the number of objects in the dataset. Moreover, the performances of these test methods using different ensemble size $M$ will be further evaluated in Section~\ref{sec:comp_Msize}.

\subsection{Comparison with Other Ensemble Clustering Methods}

In this section, we compare the proposed ECPCS-HC and ECPCS-MC methods against the baseline ensemble clustering methods on the ten benchmark datasets. For the experiment on each benchmark dataset, two criteria are adopted in terms of the number of clusters, that is, true-$k$ and best-$k$. In the true-$k$ criterion, the true number of classes in a dataset is used as the cluster number for all the test methods. In the best-$k$ criterion, the cluster number that leads to the best performance is used for each test method.

Table~\ref{table:compare_ensembles_nmi} reports the average NMI scores (over 20 runs) by different ensemble clustering methods. As shown in Table~\ref{table:compare_ensembles_nmi}, our ECPCS-HC method obtains the best performance w.r.t. NMI in terms of both true-$k$ and best-$k$ on the \emph{BC}, \emph{Ecoli}, \emph{Wine}, and \emph{Yeast} datasets, whereas ECPCS-MC achieves the best NMI scores in terms of both true-$k$ and best-$k$ on the \emph{CTG}, \emph{LR}, \emph{LS}, and \emph{PD} datasets. Note that, with two comparisons (w.r.t. true-$k$ and best-$k$ respectively) on each of ten datasets, there are totally twenty comparisons in Table~\ref{table:compare_ensembles_nmi}. As shown in Fig.~\ref{fig:rankTop1Top3_nmi_1}, our ECPCS-HC and ECPCS-MC methods are ranked in the first position in ten and nine comparisons, respectively, out of the totally twenty comparisons, while the third best method (i.e., PTGP) is ranked in the first position in only two comparisons. Similarly, as shown in Fig.~\ref{fig:rankTop1Top3_nmi_2}, ECPCS-HC and ECPCS-MC are ranked in the top 3 in seventeen and twenty comparisons, respectively, out of the totally twenty comparisons, while the third best method PTGP is only able to be ranked in the top 3 in eight comparisons.

\begin{figure*}[!t]
\begin{center}
{\subfigure[]
{\includegraphics[width=0.895\columnwidth]{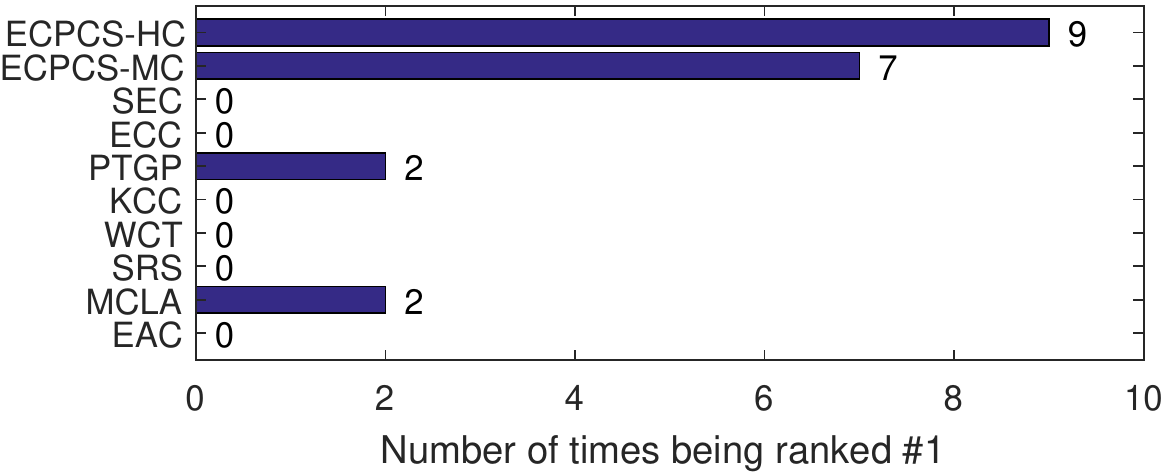}\label{fig:rankTop1Top3_ari_1}}}
{\subfigure[]
{\includegraphics[width=0.895\columnwidth]{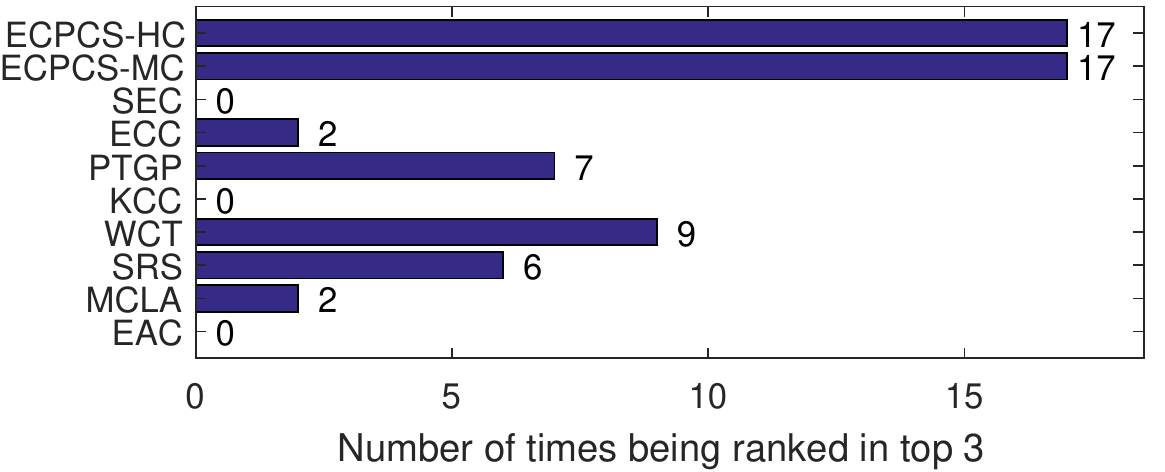}\label{fig:rankTop1Top3_ari_2}}}
\caption{The number of times that each method is ranked (a) in the first position and (b) in the top three w.r.t. Table~\ref{table:compare_ensembles_ari}.}
\label{fig:rankTop1Top3_ari}
\end{center}
\end{figure*}

Table~\ref{table:compare_ensembles_ari} reports the average ARI scores (over 20 runs) by different ensemble clustering methods. As shown in Table~\ref{table:compare_ensembles_ari}, the highest ARI scores are achieved by either ECPCS-HC or ECPCS-MC in sixteen comparisons out of the totally twenty comparisons. Specifically, as shown in Fig.~\ref{fig:rankTop1Top3_ari_1}, w.r.t. the average ARI scores, ECPCS-HC and ECPCS-MC are ranked in the first position in nine and seven comparisons, respectively, out of the totally twenty comparisons, while the third best methods is ranked in the first position in only two comparisons. As shown in Fig.~\ref{fig:rankTop1Top3_ari_2}, both ECPCS-HC and ECPCS-MC are ranked in the top 3 in seventeen comparisons out of the totally twenty comparisons, while the third best method WCT is ranked in the top 3 in only nine comparisons.

\begin{table*}[!th]
\centering
\caption{Average ARI($\%$) scores (over 20 runs) by different ensemble clustering methods. The best three scores in each comparison are highlighted in \textbf{bold}, while the best one in \textbf{[bold and brackets]}.}
\label{table:compare_ensembles_ari}
\begin{tabular}{|m{1.16cm}<{\centering}|m{0.78cm}<{\centering}|m{1.09cm}<{\centering}m{1.15cm}<{\centering}m{1.09cm}<{\centering}m{1.09cm}<{\centering}m{1.09cm}<{\centering}m{1.15cm}<{\centering}m{1.09cm}<{\centering}m{1.25cm}<{\centering}|m{1.39065cm}<{\centering}m{1.344cm}<{\centering}|}
\hline
\multicolumn{1}{|c}{\emph{Dataset}} &\multicolumn{1}{c|}{} &EAC &MCLA &SRS &WCT &KCC &PTGP &ECC &SEC &ECPCS-MC &ECPCS-HC\\
\hline
\multirow{2}{*}{\emph{BC}}	&True-$k$	&83.11$_{\pm5.29}$	&87.09$_{\pm1.57}$	&82.92$_{\pm4.19}$	&85.55$_{\pm3.73}$	&84.42$_{\pm14.33}$	&85.70$_{\pm2.90}$	&\textbf{87.68}$_{\pm1.27}$	&46.48$_{\pm35.22}$	&\textbf{87.20}$_{\pm1.06}$	&[\textbf{87.81}$_{\pm2.25}$]\\
\cline{2-12}
	&Best-$k$	&84.95$_{\pm3.36}$	&87.09$_{\pm1.57}$	&83.19$_{\pm3.67}$	&\textbf{87.58}$_{\pm1.38}$	&85.50$_{\pm8.99}$	&85.70$_{\pm2.90}$	&\textbf{87.68}$_{\pm1.27}$	&62.21$_{\pm19.35}$	&87.20$_{\pm1.06}$	&[\textbf{88.55}$_{\pm0.84}$]\\
\hline
\multirow{2}{*}{\emph{CTG}}	&True-$k$	&12.23$_{\pm0.81}$	&11.55$_{\pm0.74}$	&12.45$_{\pm0.76}$	&\textbf{12.66}$_{\pm1.02}$	&10.64$_{\pm1.39}$	&11.93$_{\pm0.96}$	&11.10$_{\pm1.06}$	&11.58$_{\pm1.54}$	&[\textbf{13.05}$_{\pm0.90}$]	&\textbf{12.68}$_{\pm1.19}$\\
\cline{2-12}
	&Best-$k$	&12.95$_{\pm1.20}$	&13.18$_{\pm0.68}$	&13.13$_{\pm1.17}$	&13.40$_{\pm1.12}$	&11.69$_{\pm0.73}$	&\textbf{13.97}$_{\pm1.11}$	&11.69$_{\pm0.75}$	&12.24$_{\pm0.95}$	&[\textbf{15.58}$_{\pm0.99}$]	&\textbf{13.79}$_{\pm0.94}$\\
\hline
\multirow{2}{*}{\emph{Ecoli}}	&True-$k$	&49.15$_{\pm5.73}$	&35.37$_{\pm4.00}$	&45.58$_{\pm5.24}$	&\textbf{57.39}$_{\pm8.79}$	&34.87$_{\pm3.89}$	&35.94$_{\pm4.16}$	&37.60$_{\pm4.16}$	&39.94$_{\pm7.64}$	&\textbf{51.44}$_{\pm2.94}$	&[\textbf{75.75}$_{\pm5.35}$]\\
\cline{2-12}
	&Best-$k$	&74.43$_{\pm2.52}$	&45.08$_{\pm2.01}$	&74.55$_{\pm1.89}$	&\textbf{76.78}$_{\pm1.65}$	&46.88$_{\pm8.17}$	&69.81$_{\pm3.38}$	&52.93$_{\pm8.89}$	&46.53$_{\pm7.87}$	&\textbf{75.44}$_{\pm1.64}$	&[\textbf{77.43}$_{\pm1.10}$]\\
\hline
\multirow{2}{*}{\emph{Gisette}}	&True-$k$	&28.10$_{\pm17.20}$	&51.31$_{\pm15.00}$	&38.99$_{\pm13.63}$	&43.65$_{\pm11.32}$	&19.54$_{\pm14.41}$	&[\textbf{58.56}$_{\pm2.09}$]	&34.63$_{\pm11.49}$	&10.98$_{\pm8.86}$	&\textbf{57.61}$_{\pm2.38}$	&\textbf{47.75}$_{\pm9.47}$\\
\cline{2-12}
	&Best-$k$	&35.42$_{\pm10.97}$	&52.95$_{\pm10.03}$	&42.01$_{\pm9.78}$	&45.15$_{\pm8.56}$	&25.58$_{\pm9.78}$	&[\textbf{58.56}$_{\pm2.09}$]	&36.14$_{\pm8.57}$	&18.73$_{\pm9.25}$	&\textbf{57.61}$_{\pm2.38}$	&\textbf{48.79}$_{\pm7.30}$\\
\hline
\multirow{2}{*}{\emph{LR}}	&True-$k$	&14.97$_{\pm0.76}$	&[\textbf{17.71}$_{\pm1.27}$]	&15.22$_{\pm1.04}$	&14.69$_{\pm0.90}$	&14.02$_{\pm1.04}$	&\textbf{16.16}$_{\pm1.36}$	&14.29$_{\pm0.66}$	&11.74$_{\pm1.79}$	&\textbf{15.44}$_{\pm0.73}$	&15.28$_{\pm0.78}$\\
\cline{2-12}
	&Best-$k$	&16.73$_{\pm0.62}$	&[\textbf{18.44}$_{\pm0.88}$]	&\textbf{17.85}$_{\pm0.70}$	&17.07$_{\pm0.55}$	&16.49$_{\pm0.80}$	&17.55$_{\pm0.73}$	&16.88$_{\pm0.72}$	&16.12$_{\pm1.29}$	&16.77$_{\pm0.39}$	&\textbf{17.68}$_{\pm0.81}$\\
\hline
\multirow{2}{*}{\emph{LS}}	&True-$k$	&56.07$_{\pm6.52}$	&46.27$_{\pm4.90}$	&57.09$_{\pm5.95}$	&\textbf{60.07}$_{\pm5.68}$	&36.28$_{\pm5.60}$	&52.68$_{\pm2.88}$	&40.24$_{\pm5.89}$	&26.57$_{\pm10.16}$	&\textbf{61.49}$_{\pm5.25}$	&[\textbf{61.62}$_{\pm5.11}$]\\
\cline{2-12}
	&Best-$k$	&60.72$_{\pm4.17}$	&52.23$_{\pm5.37}$	&62.40$_{\pm3.61}$	&\textbf{63.07}$_{\pm3.42}$	&41.29$_{\pm3.76}$	&60.76$_{\pm3.28}$	&45.87$_{\pm4.79}$	&37.10$_{\pm5.62}$	&[\textbf{65.43}$_{\pm2.60}$]	&\textbf{64.77}$_{\pm3.02}$\\
\hline
\multirow{2}{*}{\emph{MNIST}}	&True-$k$	&49.53$_{\pm2.73}$	&46.54$_{\pm5.24}$	&\textbf{51.62}$_{\pm2.45}$	&51.17$_{\pm2.06}$	&36.23$_{\pm4.11}$	&\textbf{52.88}$_{\pm4.27}$	&35.27$_{\pm3.73}$	&26.99$_{\pm6.46}$	&[\textbf{53.17}$_{\pm3.13}$]	&49.61$_{\pm1.58}$\\
\cline{2-12}
	&Best-$k$	&51.55$_{\pm2.53}$	&47.64$_{\pm4.30}$	&\textbf{54.01}$_{\pm2.12}$	&52.81$_{\pm2.30}$	&42.08$_{\pm2.61}$	&\textbf{55.43}$_{\pm3.01}$	&41.56$_{\pm1.78}$	&41.37$_{\pm2.89}$	&[\textbf{55.59}$_{\pm2.66}$]	&53.08$_{\pm2.30}$\\
\hline
\multirow{2}{*}{\emph{PD}}	&True-$k$	&62.21$_{\pm3.75}$	&58.29$_{\pm5.71}$	&63.23$_{\pm4.22}$	&62.85$_{\pm5.21}$	&43.79$_{\pm6.21}$	&\textbf{63.38}$_{\pm5.39}$	&45.09$_{\pm5.11}$	&29.90$_{\pm9.99}$	&[\textbf{65.42}$_{\pm4.23}$]	&\textbf{63.54}$_{\pm5.22}$\\
\cline{2-12}
	&Best-$k$	&71.13$_{\pm2.06}$	&60.72$_{\pm4.20}$	&\textbf{73.80}$_{\pm0.87}$	&\textbf{73.68}$_{\pm1.09}$	&56.88$_{\pm3.24}$	&72.40$_{\pm2.24}$	&56.53$_{\pm3.05}$	&54.77$_{\pm3.78}$	&73.10$_{\pm0.99}$	&[\textbf{74.61}$_{\pm0.98}$]\\
\hline
\multirow{2}{*}{\emph{Wine}}	&True-$k$	&89.56$_{\pm2.40}$	&84.50$_{\pm3.35}$	&\textbf{90.78}$_{\pm2.82}$	&90.26$_{\pm3.04}$	&88.18$_{\pm4.09}$	&90.02$_{\pm2.37}$	&84.76$_{\pm10.11}$	&88.47$_{\pm5.75}$	&\textbf{90.62}$_{\pm2.25}$	&[\textbf{91.29}$_{\pm2.96}$]\\
\cline{2-12}
	&Best-$k$	&89.76$_{\pm2.19}$	&84.50$_{\pm3.35}$	&\textbf{90.96}$_{\pm2.60}$	&90.59$_{\pm2.72}$	&88.28$_{\pm3.56}$	&90.02$_{\pm2.37}$	&86.52$_{\pm5.99}$	&88.83$_{\pm4.31}$	&\textbf{90.66}$_{\pm2.21}$	&[\textbf{91.70}$_{\pm2.53}$]\\
\hline
\multirow{2}{*}{\emph{Yeast}}	&True-$k$	&16.38$_{\pm1.58}$	&12.32$_{\pm1.03}$	&16.32$_{\pm1.31}$	&\textbf{19.01}$_{\pm1.74}$	&11.89$_{\pm2.59}$	&13.36$_{\pm1.40}$	&9.89$_{\pm0.75}$	&11.90$_{\pm2.40}$	&\textbf{16.94}$_{\pm1.33}$	&[\textbf{20.62}$_{\pm1.51}$]\\
\cline{2-12}
	&Best-$k$	&20.46$_{\pm2.53}$	&13.77$_{\pm1.12}$	&20.21$_{\pm2.77}$	&\textbf{21.40}$_{\pm1.57}$	&13.44$_{\pm1.25}$	&18.82$_{\pm1.70}$	&14.53$_{\pm0.75}$	&14.17$_{\pm1.67}$	&[\textbf{21.57}$_{\pm1.66}$]	&\textbf{21.48}$_{\pm1.19}$\\
\hline
\hline
\multirow{2}{*}{Avg. score}	&True-$k$	&46.13	&45.10	&47.42	&49.73	&37.99	&48.06	&40.05	&30.45	&51.24	&52.60\\
\cline{2-12}
	&Best-$k$	&51.81	&47.56	&53.21	&54.15	&42.81	&54.30	&45.03	&39.21	&55.90	&55.19\\
\hline
\hline
\multirow{2}{*}{Avg. rank}	&True-$k$	&5.80	&6.30	&4.70	&4.10	&8.70	&4.40	&7.90	&8.70	&2.20	&2.20\\
\cline{2-12}
	&Best-$k$	&6.40	&6.40	&4.30	&3.70	&8.50	&4.20	&7.40	&9.10	&2.70	&2.20\\
\hline
\end{tabular}
\end{table*}

Additionally, the summary statistics (i.e., average score and average rank) of the experimental results are also provided in the bottom rows of Tables~\ref{table:compare_ensembles_nmi} and \ref{table:compare_ensembles_ari}. The average score is computed by averaging the NMI (or ARI) scores of each method across the ten benchmark datsets, whereas the average rank is obtained by averaging the ranking positions of each method across the ten benchmark datasets. As shown in Table~\ref{table:compare_ensembles_nmi}, in terms of true-$k$, our ECPCS-HC method achieves the highest average NMI($\%$) score of 57.23, across the ten datasets, while ECPCS-MC achieves the second highest average score of 57.03. In terms of best-$k$, the highest two average NMI scores across the ten datasets are also obtained by the proposed ECPCS-MC and ECPCS-HC methods, respectively. When considering the average rank, ECPCS-MC and ECPCS-HC achieve the best and the second best average ranks of 1.90 and 2.30, respectively, in terms of true-$k$, which are significantly better than the third best method (i.e., WCT), whose average rank in terms of true-$k$ is 3.90. In terms of best-$k$, ECPCS-HC and ECPCS-MC are also the best two methods w.r.t. the average rank across the ten datasets. Besides the performance w.r.t. NMI, similar advantages can also be observed in terms of average score and average rank w.r.t. ARI (as shown in Table~\ref{table:compare_ensembles_ari}). Moreover, it is interesting to compare ECPCS-HC against EAC and to compare ECPCS-MC against MCLA. Since EAC is a classical method based on the conventional co-association matrix, the comparison between the proposed ECPCS-HC method (which typically incorporates the multi-scale cluster-level information via the ECA matrix) and the EAC method provides a straightforward view as to how the proposed ECA matrix improves the consensus performance when compared to the original co-association matrix. Specifically, the average NMI($\%$) and ARI($\%$) scores (in terms of true-$k$) of EAC are respectively 53.18 and 46.13, whereas that of ECPCS-HC are respectively 57.23 and 52.60. Similar improvements can also be observed when considering the best-$k$ situation (as shown in Tables~\ref{table:compare_ensembles_nmi} and \ref{table:compare_ensembles_ari}). Besides ECPCS-HC versus EAC, the ECPCS-MC versus MCLA comparison also provides a view as to what influence the multi-scale cluster-level information has upon the conventional meta-cluster based method. Note that both ECPCS-MC and MCLA are meta-cluster based methods, the integration of cluster-wise similarity propagation is able to bring in significant improvements for the ECPCS-MC method when compared to the classical MCLA method, as shown by their average scores and average ranks across ten datasets.
To summarize, as shown in Tables~\ref{table:compare_ensembles_nmi} and \ref{table:compare_ensembles_ari} and Figs.~\ref{fig:rankTop1Top3_nmi} and \ref{fig:rankTop1Top3_ari}, the proposed ECPCS-HC and ECPCS-MC methods exhibit overall better performances (w.r.t. NMI and ARI) than the baseline methods on the benchmark datasets.

\subsection{Robustness to Ensemble Size $M$}
\label{sec:comp_Msize}

\begin{figure*}[!th]
\begin{center}
{\subfigure[]
{\includegraphics[width=0.395\columnwidth]{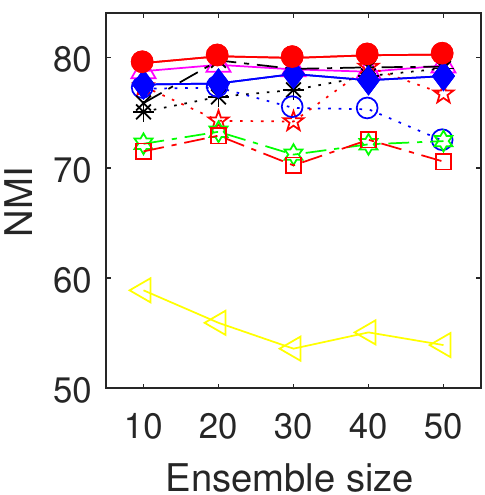}}}
{\subfigure[]
{\includegraphics[width=0.395\columnwidth]{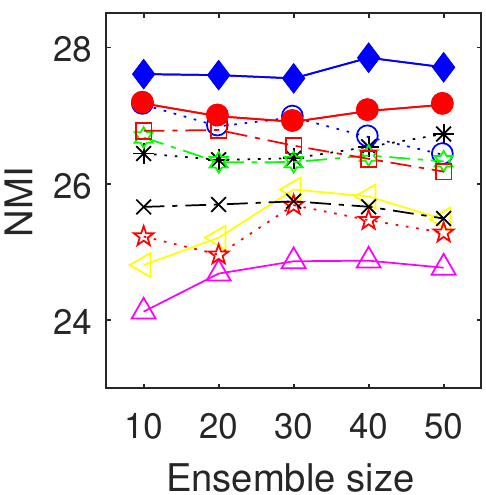}}}
{\subfigure[]
{\includegraphics[width=0.395\columnwidth]{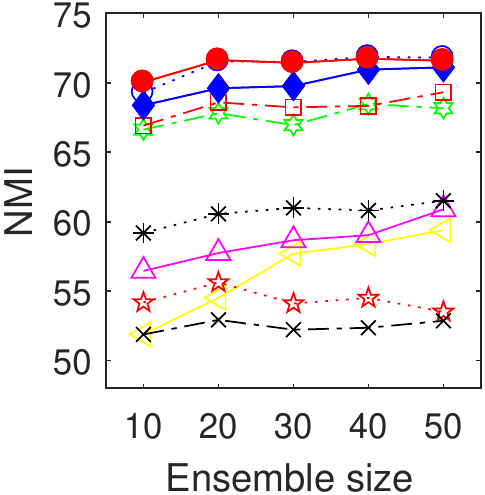}}}
{\subfigure[]
{\includegraphics[width=0.395\columnwidth]{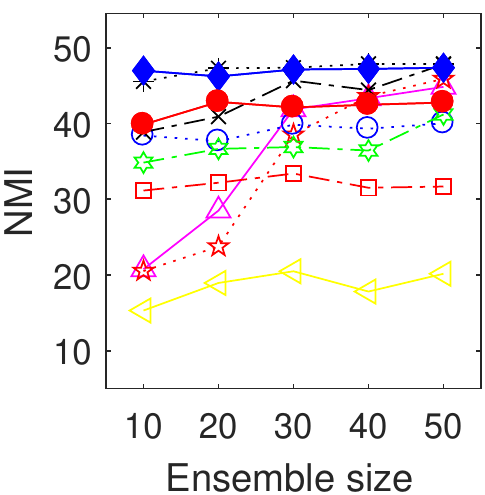}}}
{\subfigure[]
{\includegraphics[width=0.395\columnwidth]{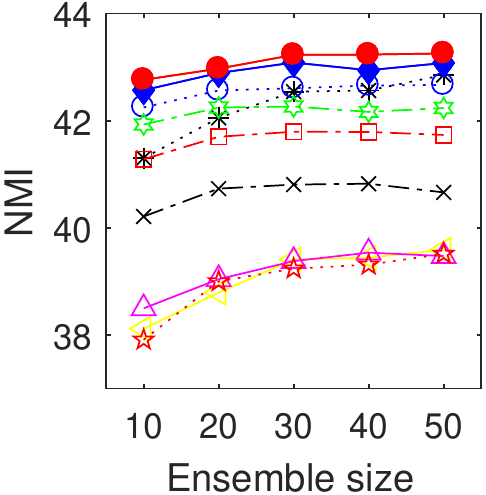}}}
{\subfigure[]
{\includegraphics[width=0.395\columnwidth]{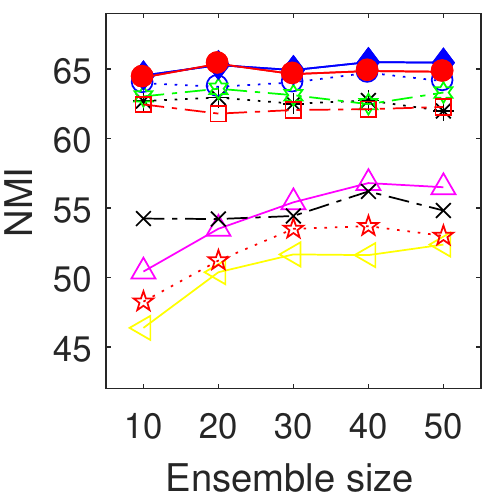}}}
{\subfigure[]
{\includegraphics[width=0.395\columnwidth]{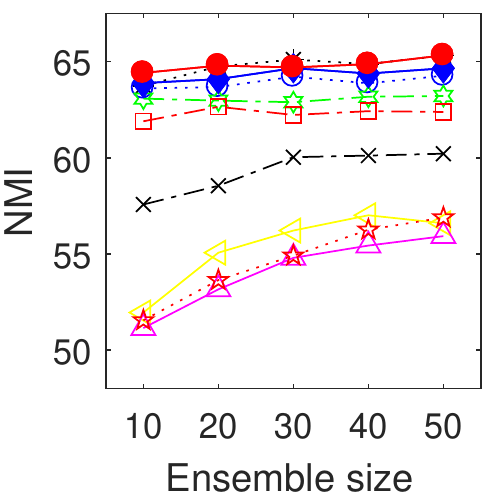}}}
{\subfigure[]
{\includegraphics[width=0.395\columnwidth]{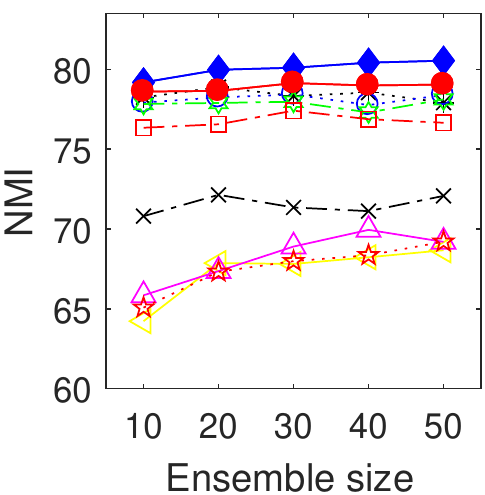}}}
{\subfigure[]
{\includegraphics[width=0.395\columnwidth]{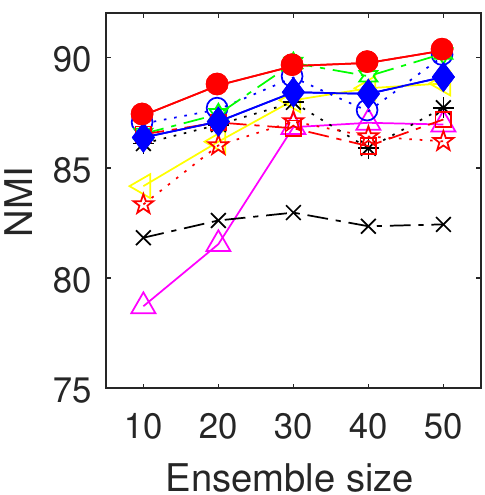}}}
{\subfigure[]
{\includegraphics[width=0.395\columnwidth]{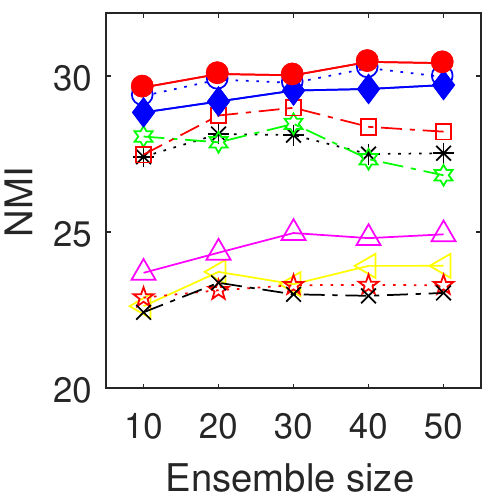}}}
{\subfigure
{\includegraphics[width=1.4\columnwidth]{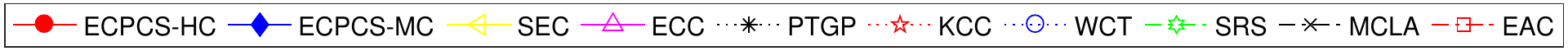}}}
\caption{Average performances w.r.t. NMI($\%$) over 20 runs by different ensemble clustering methods with varying ensemble sizes $M$. (a) \emph{BC}. (b) \emph{CTG}. (c) \emph{Ecoli}. (d) \emph{Gisette}. (e) \emph{LR}. (f) \emph{LS}. (g) \emph{MNIST}. (h) \emph{PD}. (i) \emph{Wine}. (j) \emph{Yeast}.}
\label{fig:comp_Msize_nmi}
\end{center}
\end{figure*}

\begin{figure*}[!th]
\begin{center}
{\subfigure[]
{\includegraphics[width=0.395\columnwidth]{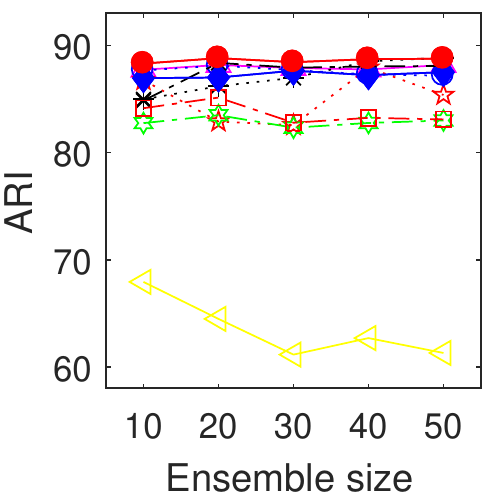}}}
{\subfigure[]
{\includegraphics[width=0.395\columnwidth]{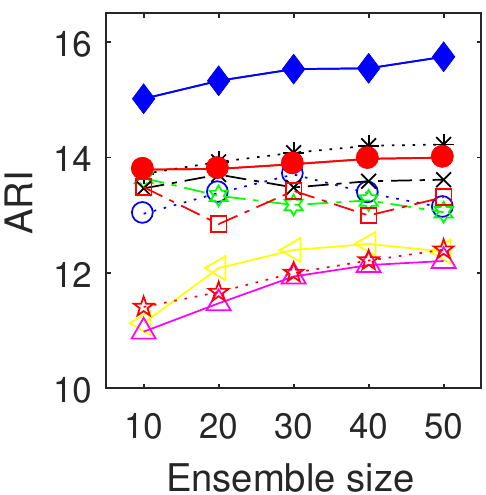}}}
{\subfigure[]
{\includegraphics[width=0.395\columnwidth]{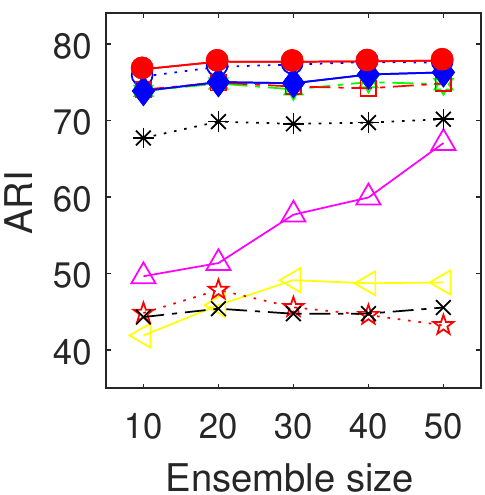}}}
{\subfigure[]
{\includegraphics[width=0.395\columnwidth]{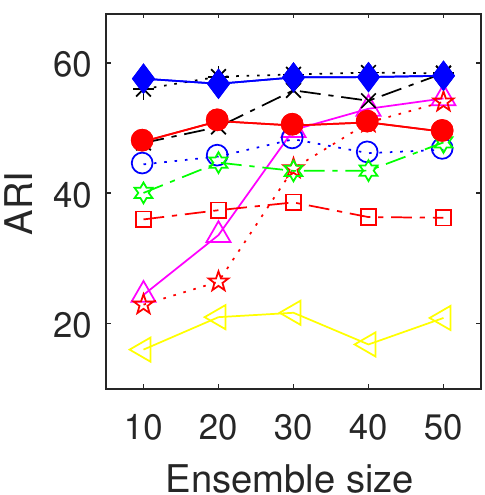}}}
{\subfigure[]
{\includegraphics[width=0.395\columnwidth]{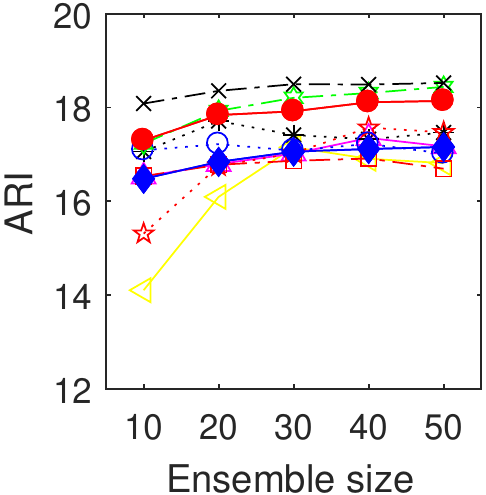}}}
{\subfigure[]
{\includegraphics[width=0.395\columnwidth]{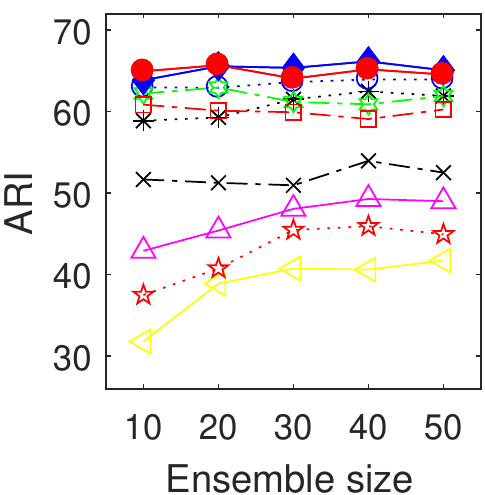}}}
{\subfigure[]
{\includegraphics[width=0.395\columnwidth]{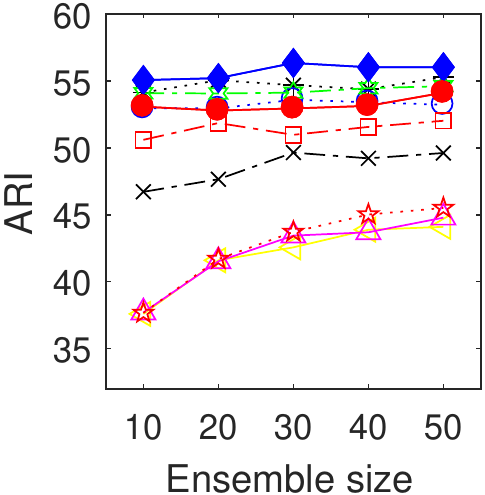}}}
{\subfigure[]
{\includegraphics[width=0.395\columnwidth]{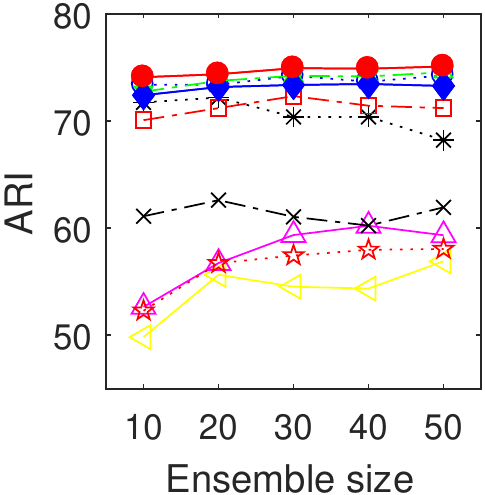}}}
{\subfigure[]
{\includegraphics[width=0.395\columnwidth]{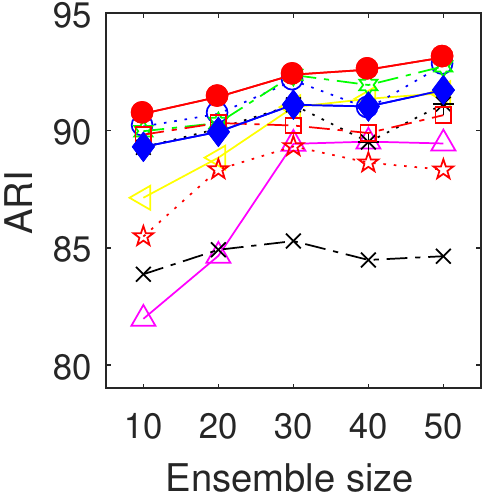}}}
{\subfigure[]
{\includegraphics[width=0.395\columnwidth]{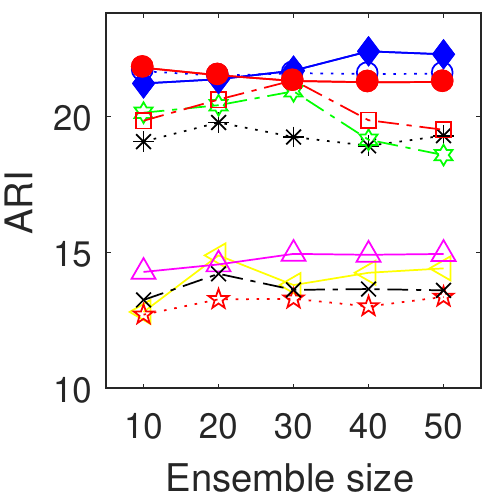}}}
{\subfigure
{\includegraphics[width=1.4\columnwidth]{Figures_cmpEnSize_legend}}}
\caption{Average performances w.r.t. ARI($\%$) over 20 runs by different ensemble clustering methods with varying ensemble sizes $M$. (a) \emph{BC}. (b) \emph{CTG}. (c) \emph{Ecoli}. (d) \emph{Gisette}. (e) \emph{LR}. (f) \emph{LS}. (g) \emph{MNIST}. (h) \emph{PD}. (i) \emph{Wine}. (j) \emph{Yeast}.}
\label{fig:comp_Msize_ari}
\end{center}
\end{figure*}

In this section, we evaluate the performances of the proposed methods and the baseline methods using different ensemble sizes $M$. As shown in Fig.~\ref{fig:comp_Msize_nmi}, ECPCS-HC obtains the best performance w.r.t. NMI on the \emph{BC}, \emph{Ecoli}, \emph{LR}, \emph{wine}, and \emph{yeast} datasets, whereas ECPCS-MC obtains the best performance on the \emph{CTG} and \emph{PD} datasets, as the ensemble size goes from 10 to 50. Similarly, as shown in Fig.~\ref{fig:comp_Msize_ari}, ECPCS-HC obtains the best performance (w.r.t. ARI) on the \emph{BC}, \emph{Ecoli}, \emph{PD}, and \emph{Wine} datasets, whereas ECPCS-MC obtains the best performance (w.r.t. ARI) on the \emph{CTG} and \emph{MNIST} datasets, with varying ensemble sizes $M$. Although the MCLA method shows better ARI scores than the proposed methods on the \emph{LR} dataset, yet on all of the other nine datasets our methods consistently outperform MCLA with different ensemble sizes. As can be seen in Figs.~\ref{fig:comp_Msize_nmi} and \ref{fig:comp_Msize_ari}, the proposed ECPCS-HC and ECPCS-MC methods exhibit overall the best performance w.r.t. NMI and ARI on the benchmark datasets.

\subsection{Sensitivity of Parameter $t$}
\label{sec:sensitivity_t}

In this section, we evaluate the performances of the proposed ECPCS-HC and ECPCS-MC methods with varying parameter $t$.

Table~\ref{table:cmpPara_NMI} reports the average NMI scores (over 20 runs) of ECPCS-HC and ECPCS-MC when the parameter $t$ takes different values. Note that the parameter $t$ controls the number of steps of the random walkers during the propagation of cluster-wise similarities (as described in Section~\ref{sec:propagation_of_CSG}). As shown in Table~\ref{table:cmpPara_NMI}, the proposed methods yield consistently good performances (w.r.t. NMI) with varying parameter $t$. Generally, using a larger parameter $t$ (e.g., larger than 10) can lead to better clustering results than using a very small one, which is probably due to the fact that a random walker with adequate number of steps can better capture the multi-scale structure information of the graph. Also, the performances (w.r.t. ARI) by the proposed ECPCS-HC and ECPCS-MC methods are reported in Table~\ref{table:cmpPara_ARI}. From the experimental results in Tables~\ref{table:cmpPara_NMI} and \ref{table:cmpPara_ARI}, it can be observed that the proposed methods exhibit robust consensus clustering performances with different values of the parameter.

\begin{table*}[!t]
\centering
\caption{Average performance w.r.t. NMI($\%$) over 20 runs by our ECPCS-HC and ECPCS-MC methods using varying parameter $t$.}
\label{table:cmpPara_NMI}
\begin{tabular}{|m{0.99cm}<{\centering}|m{0.801cm}<{\centering}m{0.701cm}<{\centering}m{0.701cm}<{\centering}m{0.701cm}<{\centering}m{0.701cm}<{\centering}m{0.701cm}<{\centering}m{0.721cm}<{\centering}|m{0.801cm}<{\centering}m{0.701cm}<{\centering}m{0.701cm}<{\centering}m{0.701cm}<{\centering}m{0.701cm}<{\centering}m{0.701cm}<{\centering}m{0.721cm}<{\centering}|}
\hline
\multirow{2}{*}{Dataset} &\multicolumn{7}{c|}{ECPCS-HC} &\multicolumn{7}{c|}{ECPCS-MC}\\
\cline{2-15}
&$t=1$ &2 &4 &8 &16 &32 &64 &$t=1$ &2 &4 &8 &16 &32 &64\\
\hline
\emph{BC}	&75.30	&76.28	&76.86	&78.23	&79.16	&79.69	&79.99	&77.19	&77.48	&77.80	&77.82	&77.80	&77.81	&77.74\\
\hline
\emph{CTG}	&26.65	&26.69	&26.82	&26.90	&26.99	&26.97	&27.10	&27.45	&27.63	&27.63	&27.66	&27.66	&27.69	&27.73\\
\hline
\emph{Ecoli}	&70.61	&70.98	&71.44	&71.66	&71.49	&71.34	&71.26	&69.63	&69.75	&70.11	&70.13	&70.09	&70.04	&70.08\\
\hline
\emph{Gisette}	&37.58	&38.60	&40.34	&41.01	&40.95	&41.00	&39.49	&46.51	&46.77	&46.99	&47.04	&46.97	&46.88	&46.78\\
\hline
\emph{LR}	&42.42	&42.35	&42.45	&42.50	&42.75	&43.20	&43.56	&42.36	&42.75	&42.96	&42.90	&42.85	&42.78	&42.77\\
\hline
\emph{LS}	&63.43	&63.78	&64.22	&64.50	&64.95	&64.93	&65.00	&65.19	&65.31	&65.32	&65.29	&65.18	&65.16	&65.03\\
\hline
\emph{MNIST}	&63.36	&63.43	&63.94	&64.32	&64.88	&64.82	&64.50	&63.63	&63.75	&63.90	&64.01	&64.30	&64.34	&64.22\\
\hline
\emph{PD}	&77.98	&78.35	&78.53	&78.80	&78.73	&78.78	&78.49	&79.66	&79.81	&79.80	&79.84	&79.77	&79.70	&79.80\\
\hline
\emph{Wine}	&87.22	&88.24	&88.52	&88.65	&88.98	&88.95	&89.06	&87.41	&87.69	&87.91	&87.87	&87.91	&87.91	&87.86\\
\hline
\emph{Yeast}	&29.25	&29.35	&29.48	&29.91	&30.04	&30.15	&29.97	&28.73	&28.89	&29.03	&29.20	&29.22	&29.36	&29.44\\
\hline
\end{tabular}
\end{table*}

\begin{table*}[!t]
\centering
\caption{Average performance w.r.t. ARI($\%$) over 20 runs by our ECPCS-HC and ECPCS-MC methods using varying parameter $t$.}
\label{table:cmpPara_ARI}
\begin{tabular}{|m{0.99cm}<{\centering}|m{0.801cm}<{\centering}m{0.701cm}<{\centering}m{0.701cm}<{\centering}m{0.701cm}<{\centering}m{0.701cm}<{\centering}m{0.701cm}<{\centering}m{0.721cm}<{\centering}|m{0.801cm}<{\centering}m{0.701cm}<{\centering}m{0.701cm}<{\centering}m{0.701cm}<{\centering}m{0.701cm}<{\centering}m{0.701cm}<{\centering}m{0.721cm}<{\centering}|}
\hline
\multirow{2}{*}{Dataset} &\multicolumn{7}{c|}{ECPCS-HC} &\multicolumn{7}{c|}{ECPCS-MC}\\
\cline{2-15}
&$t=1$ &2 &4 &8 &16 &32 &64 &$t=1$ &2 &4 &8 &16 &32 &64\\
\hline
\emph{BC}	&86.75	&87.34	&87.56	&88.12	&88.52	&88.67	&88.74	&87.03	&87.12	&87.14	&87.15	&87.14	&87.14	&87.09\\
\hline
\emph{CTG}	&13.39	&13.47	&13.65	&13.65	&13.49	&13.49	&13.61	&15.52	&15.56	&15.73	&15.66	&15.62	&15.65	&15.62\\
\hline
\emph{Ecoli}	&76.40	&76.75	&77.17	&77.41	&77.41	&77.35	&77.33	&75.34	&75.44	&75.67	&75.59	&75.57	&75.53	&75.63\\
\hline
\emph{Gisette}	&43.75	&45.15	&47.22	&48.81	&48.76	&48.82	&47.18	&57.09	&57.36	&57.58	&57.63	&57.57	&57.47	&57.37\\
\hline
\emph{LR}	&17.05	&17.02	&17.07	&17.24	&17.43	&17.35	&16.93	&16.55	&16.59	&16.63	&16.62	&16.65	&16.75	&16.75\\
\hline
\emph{LS}	&62.89	&63.11	&64.16	&64.51	&64.82	&64.54	&63.94	&65.89	&66.06	&66.29	&66.04	&65.74	&65.23	&64.89\\
\hline
\emph{MNIST}	&52.37	&52.61	&53.12	&53.08	&53.20	&52.67	&52.16	&53.63	&54.00	&54.46	&54.83	&55.45	&55.45	&55.34\\
\hline
\emph{PD}	&73.03	&73.69	&74.11	&74.47	&74.59	&74.68	&74.82	&73.03	&73.09	&73.17	&73.12	&73.11	&73.20	&73.20\\
\hline
\emph{Wine}	&90.57	&91.36	&91.54	&91.63	&91.81	&91.78	&91.99	&90.69	&90.69	&90.70	&90.67	&90.70	&90.69	&90.66\\
\hline
\emph{Yeast}	&20.99	&21.15	&21.46	&21.72	&21.61	&21.41	&21.04	&21.12	&21.19	&21.29	&21.34	&21.39	&21.70	&21.89\\
\hline
\end{tabular}
\end{table*}

\subsection{Execution Time}

\begin{figure}[!t]
\begin{center}
{
{\includegraphics[width=0.91\columnwidth]{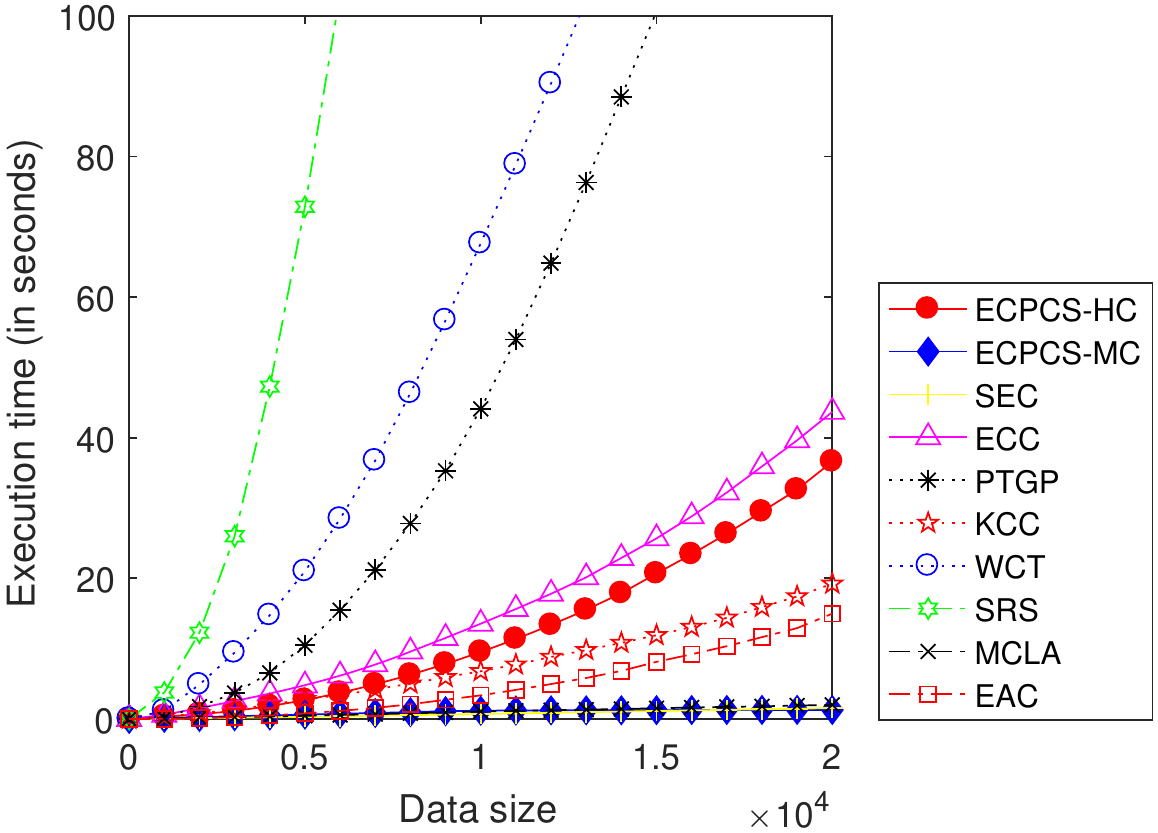}}}
\caption{Execution times of different ensemble clustering methods on the \emph{LR} datasets with the data size varying from 0 to 20,000.}
\label{fig:compTimeAll}
\end{center}
\end{figure}

In this section, we evaluate the execution times of different ensemble clustering methods. The experiments are conducted on the \emph{LR} dataset with the data size varying from 0 to 20,000.  As shown in Fig.~\ref{fig:compTimeAll}, ECPCS-MC is the fastest method, which requires 1.28s to process the entire \emph{LR} dataset with 20,000 objects, while SEC and MCLA are the second and third fastest ones, which requires 1.56s and 2.00s, respectively, to process the entire \emph{LR} dataset. The time efficiency of ECPCS-HC is comparable to that of the ECC method, and is better than the PTGP, WCT, and SRS methods. To summarize, the proposed ECPCS-MC and ECPCS-HC methods consistently outperform the baseline methods in terms of clustering quality (as shown in Tables~\ref{table:compare_ensembles_nmi} and \ref{table:compare_ensembles_ari} and Figs.~\ref{fig:rankTop1Top3_nmi}, \ref{fig:rankTop1Top3_ari}, \ref{fig:comp_Msize_nmi}, and \ref{fig:comp_Msize_ari}) while exhibiting competitive time efficiency (as shown in Fig.~\ref{fig:compTimeAll}).

All experiments were conducted in MATLAB 2016b on a PC with an Intel i7-6700K CPU and 64GB of RAM.

\section{Conclusion}
\label{sec:conclusion}

In this paper, we propose a new ensemble clustering approach based on fast propagation of cluster-wise similarities via random walks. By treating the base clusters as nodes and using the Jaccard coefficient to build weighted edges, a cluster similarity graph is first constructed. With a new transition probability matrix defined on the graph, the random walk process is performed with each node treated as a starting node. Then, a new cluster-wise similarity matrix can be derived from the original graph by investigating the propagating trajectories of the random walkers starting from different nodes (i.e., clusters). Then, we construct an ECA matrix by mapping the new cluster-wise similarity from the cluster-level to the object-level, and propose two novel consensus functions, i.e., ECPCS-HC and ECPCS-MC, to achieve the final consensus clustering result. Extensive experiments are conducted on ten real-world datasets, which have shown the advantage of our ensemble clustering approach over the state-of-the-art in terms of both clustering quality and efficiency.

\ifCLASSOPTIONcompsoc
  \section*{Acknowledgments}
\else
  \section*{Acknowledgment}
\fi

The authors would like to thank the anonymous reviewers for their constructive comments and suggestions that help enhance the paper significantly.

Our MATLAB source code is available for download at: www.researchgate.net/publication/328581758.

\ifCLASSOPTIONcaptionsoff
  \newpage
\fi

\bibliographystyle{IEEEtran}
\bibliography{tsmcs_2018}

\begin{thebibliography}{10}
\providecommand{\url}[1]{#1}
\csname url@samestyle\endcsname
\providecommand{\newblock}{\relax}
\providecommand{\bibinfo}[2]{#2}
\providecommand{\BIBentrySTDinterwordspacing}{\spaceskip=0pt\relax}
\providecommand{\BIBentryALTinterwordstretchfactor}{4}
\providecommand{\BIBentryALTinterwordspacing}{\spaceskip=\fontdimen2\font plus
\BIBentryALTinterwordstretchfactor\fontdimen3\font minus
  \fontdimen4\font\relax}
\providecommand{\BIBforeignlanguage}[2]{{%
\expandafter\ifx\csname l@#1\endcsname\relax
\typeout{** WARNING: IEEEtran.bst: No hyphenation pattern has been}%
\typeout{** loaded for the language `#1'. Using the pattern for}%
\typeout{** the default language instead.}%
\else
\language=\csname l@#1\endcsname
\fi
#2}}
\providecommand{\BIBdecl}{\relax}
\BIBdecl

\bibitem{frey07_ap}
B.~J. Frey and D.~Dueck, ``Clustering by passing messages between data
  points,'' \emph{Science}, vol. 315, pp. 972--976, 2007.

\bibitem{das08_tsmca}
S.~Das, A.~Abraham, and A.~Konar, ``Automatic clustering using an improved
  differential evolution algorithm,'' \emph{IEEE Transactions on Systems, Man,
  and Cybernetics - Part A: Systems and Humans}, vol.~38, no.~1, pp. 218--237,
  2008.

\bibitem{meap13}
C.-D. Wang, J.-H. Lai, C.~Y. Suen, and J.-Y. Zhu, ``Multi-exemplar affinity
  propagation,'' \emph{IEEE Transactions on Pattern Analysis and Machine
  Intelligence}, vol.~35, no.~9, pp. 2223--2237, 2013.

\bibitem{svstream13}
C.-D. Wang, J.-H. Lai, D.~Huang, and W.-S. Zheng, ``{SVS}tream: A support
  vector based algorithm for clustering data streams,'' \emph{IEEE Transactions
  on Knowledge and Data Engineering}, vol.~25, no.~6, pp. 1410--1424, 2013.

\bibitem{yang15}
Y.~Yang, Z.~Ma, Y.~Yang, F.~Nie, and H.~T. Shen, ``Multitask spectral
  clustering by exploring intertask correlation,'' \emph{IEEE Transactions on
  Cybernetics}, vol.~45, no.~5, pp. 1083--1094, 2015.

\bibitem{wang16_tkde}
C.-D. Wang, J.-H. Lai, and P.~S. Yu, ``Multi-view clustering based on belief
  propagation,'' \emph{IEEE Transactions on Knowledge and Data Engineering},
  vol.~28, no.~4, pp. 1007--1021, 2016.

\bibitem{Chen18_tsmcs}
Y.~Chen, S.~Tang, S.~Pei, C.~Wang, J.~Du, and N.~Xiong, ``D{H}eat: A density
  heat-based algorithm for clustering with effective radius,'' \emph{IEEE
  Transactions on Systems, Man, and Cybernetics: Systems}, vol.~48, no.~4, pp.
  649--660, 2018.

\bibitem{Zhang18_tsmcs}
Y.~Zhang, F.~L. Chung, and S.~Wang, ``Fast reduced set-based exemplar finding
  and cluster assignment,'' \emph{IEEE Transactions on Systems, Man, and
  Cybernetics: Systems, in press}, 2018.

\bibitem{He18_tsmcs}
H.~He and Y.~Tan, ``Pattern clustering of hysteresis time series with
  multivalued mapping using tensor decomposition,'' \emph{IEEE Transactions on
  Systems, Man, and Cybernetics: Systems}, vol.~48, no.~6, pp. 993--1004, 2018.

\bibitem{wu17_Euler}
J.~S. Wu, W.~S. Zheng, J.~H. Lai, and C.~Y. Suen, ``Euler clustering on
  large-scale dataset,'' \emph{IEEE Transactions on Big Data, in press}, 2018.

\bibitem{jm00_ncut}
J.~Shi and J.~Malik, ``Normalized cuts and image segmentation,'' \emph{IEEE
  Transactions on Pattern Analysis and Machine Intelligence}, vol.~22, no.~8,
  pp. 888--905, 2000.

\bibitem{Huang16_neucom}
D.~Huang, J.-H. Lai, C.-D. Wang, and P.~C. Yuen, ``Ensembling
  over-segmentations: From weak evidence to strong segmentation,''
  \emph{Neurocomputing}, vol. 207, pp. 416--427, 2016.

\bibitem{Wang14_tsmcs}
Z.~Wang, D.~Zhang, X.~Zhou, D.~Yang, Z.~Yu, and Z.~Yu, ``Discovering and
  profiling overlapping communities in location-based social networks,''
  \emph{IEEE Transactions on Systems, Man, and Cybernetics: Systems}, vol.~44,
  no.~4, pp. 499--509, 2014.

\bibitem{neiwalk14_tkde}
C.-D. Wang, J.-H. Lai, and P.~S. Yu, ``{NEIW}alk: Community discovery in
  dynamic content-based networks,'' \emph{IEEE Transactions on Knowledge and
  Data Engineering}, vol.~26, no.~7, pp. 1734--1748, 2014.

\bibitem{rafa13_tsmcs}
D.~Rafailidis and P.~Daras, ``The {TFC} model: Tensor factorization and tag
  clustering for item recommendation in social tagging systems,'' \emph{IEEE
  Transactions on Systems, Man, and Cybernetics: Systems}, vol.~43, no.~3, pp.
  673--688, 2013.

\bibitem{symeon16_tsmcs}
P.~Symeonidis, ``Clusthosvd: Item recommendation by combining semantically
  enhanced tag clustering with tensor hosvd,'' \emph{IEEE Transactions on
  Systems, Man, and Cybernetics: Systems}, vol.~46, no.~9, pp. 1240--1251,
  2016.

\bibitem{zhao18_tsmcs}
Q.~Zhao, C.~Wang, P.~Wang, M.~Zhou, and C.~Jiang, ``A novel method on
  information recommendation via hybrid similarity,'' \emph{IEEE Transactions
  on Systems, Man, and Cybernetics: Systems}, vol.~48, no.~3, pp. 448--459,
  2018.

\bibitem{rajp14_tsmcs}
D.~G. Rajpathak and S.~Singh, ``An ontology-based text mining method to develop
  d-matrix from unstructured text,'' \emph{IEEE Transactions on Systems, Man,
  and Cybernetics: Systems}, vol.~44, no.~7, pp. 966--977, 2014.

\bibitem{jain10_survey}
A.~K. Jain, ``Data clustering: 50 years beyond $k$-means,'' \emph{Pattern
  Recognition Letters}, vol.~31, no.~8, pp. 651--666, 2010.

\bibitem{Fred05_EAC}
A.~L.~N. Fred and A.~K. Jain, ``Combining multiple clusterings using evidence
  accumulation,'' \emph{IEEE Transactions on Pattern Analysis and Machine
  Intelligence}, vol.~27, no.~6, pp. 835--850, 2005.

\bibitem{huang14_weac}
D.~Huang, J.-H. Lai, and C.-D. Wang, ``Combining multiple clusterings via crowd
  agreement estimation and multi-granularity link analysis,''
  \emph{Neurocomputing}, vol. 170, pp. 240--250, 2015.

\bibitem{Yu14_pr}
Z.~Yu, L.~Li, Y.~Gao, J.~You, J.~Liu, H.-S. Wong, and G.~Han, ``Hybrid
  clustering solution selection strategy,'' \emph{Pattern Recognition},
  vol.~47, no.~10, pp. 3362--3375, 2014.

\bibitem{wu15_TKDE}
J.~Wu, H.~Liu, H.~Xiong, J.~Cao, and J.~Chen, ``K-means-based consensus
  clustering: A unified view,'' \emph{IEEE Transactions on Knowledge and Data
  Engineering}, vol.~27, no.~1, pp. 155--169, 2015.

\bibitem{huang15_ecfg}
D.~Huang, J.~Lai, and C.-D. Wang, ``Ensemble clustering using factor graph,''
  \emph{Pattern Recognition}, vol.~50, pp. 131--142, 2016.

\bibitem{Huang16_TKDE}
D.~Huang, J.-H. Lai, and C.-D. Wang, ``Robust ensemble clustering using
  probability trajectories,'' \emph{IEEE Transactions on Knowledge and Data
  Engineering}, vol.~28, no.~5, pp. 1312--1326, 2016.

\bibitem{huang17_tcyb}
D.~Huang, C.~D. Wang, and J.~H. Lai, ``Locally weighted ensemble clustering,''
  \emph{IEEE Transactions on Cybernetics}, vol.~48, no.~5, pp. 1460--1473,
  2018.

\bibitem{Yu16_tkde_incremental}
Z.~Yu, P.~Luo, J.~You, H.~S. Wong, H.~Leung, S.~Wu, J.~Zhang, and G.~Han,
  ``Incremental semi-supervised clustering ensemble for high dimensional data
  clustering,'' \emph{IEEE Transactions on Knowledge and Data Engineering},
  vol.~28, no.~3, pp. 701--714, 2016.

\bibitem{Kang16_kbs}
Q.~Kang, S.~Liu, M.~Zhou, and S.~Li, ``A weight-incorporated similarity-based
  clustering ensemble method based on swarm intelligence,''
  \emph{Knowledge-Based Systems}, vol. 104, pp. 156--164, 2016.

\bibitem{huang17_iconip}
D.~Huang, C.-D. Wang, and J.-H. Lai, ``{LWMC}: A locally weighted
  meta-clustering algorithm for ensemble clustering,'' in \emph{Proc. of
  International Conference on Neural Information Processing (ICONIP)}, 2017,
  pp. 167--176.

\bibitem{liu17_tkde}
H.~Liu, J.~Wu, T.~Liu, D.~Tao, and Y.~Fu, ``Spectral ensemble clustering via
  weighted k-means: Theoretical and practical evidence,'' \emph{IEEE
  Transactions on Knowledge and Data Engineering}, vol.~29, no.~5, pp.
  1129--1143, 2017.

\bibitem{Yu17_tkde}
Z.~Yu, Z.~Kuang, J.~Liu, H.~Chen, J.~Zhang, J.~You, H.~S. Wong, and G.~Han,
  ``Adaptive ensembling of semi-supervised clustering solutions,'' \emph{IEEE
  Transactions on Knowledge and Data Engineering}, vol.~29, no.~8, pp.
  1577--1590, 2017.

\bibitem{yu17_tcyb}
Z.~Yu, X.~Zhu, H.~Wong, J.~You, J.~Zhang, and G.~Han, ``Distribution-based
  cluster structure selection,'' \emph{IEEE Transactions on Cybernetics},
  vol.~47, no.~11, pp. 3554--3567, 2017.

\bibitem{iam_on11_linkbased}
N.~Iam-On, T.~Boongoen, S.~Garrett, and C.~Price, ``A link-based approach to
  the cluster ensemble problem,'' \emph{IEEE Transactions on Pattern Analysis
  and Machine Intelligence}, vol.~33, no.~12, pp. 2396--2409, 2011.

\bibitem{iamon08_icds}
N.~Iam-On, T.~Boongoen, and S.~Garrett, ``Refining pairwise similarity matrix
  for cluster ensemble problem with cluster relations,'' in \emph{Proc. of
  International Conference on Discovery Science (ICDS)}, 2008, pp. 222--233.

\bibitem{strehl02}
A.~Strehl and J.~Ghosh, ``Cluster ensembles: A knowledge reuse framework for
  combining multiple partitions,'' \emph{Journal of Machine Learning Research},
  vol.~3, pp. 583--617, 2003.

\bibitem{fern04_bipartite}
X.~Z. Fern and C.~E. Brodley, ``Solving cluster ensemble problems by bipartite
  graph partitioning,'' in \emph{Proc. of International Conference on Machine
  Learning (ICML)}, 2004.

\bibitem{topchy05}
A.~Topchy, A.~K. Jain, and W.~Punch, ``Clustering ensembles: models of
  consensus and weak partitions,'' \emph{IEEE Transactions on Pattern Analysis
  and Machine Intelligence}, vol.~27, no.~12, pp. 1866--1881, 2005.

\bibitem{li07}
Y.~Li, J.~Yu, P.~Hao, and Z.~Li, ``Clustering ensembles based on normalized
  edges,'' in \emph{Proc. of Pacific-Asia Conference on Knowledge Discovery and
  Data Mining (PAKDD)}, 2007.

\bibitem{Li_WCC08}
T.~Li and C.~Ding, ``Weighted consensus clustering,'' in \emph{Proc. of SIAM
  International Conference on Data Mining (SDM)}, 2008, pp. 798--809.

\bibitem{Mimaroglu11_pr}
S.~Mimaroglu and E.~Erdil, ``Combining multiple clusterings using similarity
  graph,'' \emph{Pattern Recognition}, vol.~44, no.~3, pp. 694--703, 2011.

\bibitem{yi_icdm12}
J.~Yi, T.~Yang, R.~Jin, and A.~K. Jain, ``Robust ensemble clustering by matrix
  completion,'' in \emph{Proc. of IEEE International Conference on Data Mining
  (ICDM)}, 2012.

\bibitem{franek13_pr}
L.~Franek and X.~Jiang, ``Ensemble clustering by means of clustering embedding
  in vector spaces,'' \emph{Pattern Recognition}, vol.~47, no.~2, pp. 833--842,
  2014.

\bibitem{Zhong15_pr}
C.~Zhong, X.~Yue, Z.~Zhang, and J.~Lei, ``A clustering ensemble:
  Two-level-refined co-association matrix with path-based transformation,''
  \emph{Pattern Recognition}, vol.~48, no.~8, pp. 2699--2709, 2015.

\bibitem{liu17_bioinformatics}
H.~Liu, R.~Zhao, H.~Fang, F.~Cheng, Y.~Fu, and Y.-Y. Liu, ``Entropy-based
  consensus clustering for patient stratification,'' \emph{Bioinformatics},
  vol.~33, no.~17, pp. 2691--2698, 2017.

\bibitem{Weiszfeld09}
E.~Weiszfeld and F.~Plastria, ``On the point for which the sum of the distances
  to n given points is minimum,'' \emph{Annals of Operations Research}, vol.
  167, no.~1, pp. 7--41, 2009.

\bibitem{Kschischang_FG_SPA:01}
F.~R. Kschischang, B.~J. Frey, and H.-A. Loeliger, ``Factor graphs and the
  sum-product algorithm,'' \emph{IEEE Transactions on Information Theory},
  vol.~47, no.~2, pp. 498--519, 2001.

\bibitem{ren13_icdm}
Y.~Ren, C.~Domeniconi, G.~Zhang, and G.~Yu, ``Weighted-object ensemble
  clustering,'' in \emph{Proc. of International Conference on Data Mining
  (ICDM)}, 2013, pp. 627--636.

\bibitem{yu15_tkde}
Z.~Yu, L.~Li, J.~Liu, J.~Zhang, and G.~Han, ``Adaptive noise immune cluster
  ensemble using affinity propagation,'' \emph{IEEE Transactions on Knowledge
  and Data Engineering}, vol.~27, no.~12, pp. 3176--3189, 2015.

\bibitem{yu15_tcbb}
Z.~Yu, H.~Chen, J.~You, J.~Liu, H.~Wong, G.~Han, and L.~Li, ``Adaptive fuzzy
  consensus clustering framework for clustering analysis of cancer data,''
  \emph{IEEE/ACM Transactions on Computational Biology and Bioinformatics},
  vol.~12, no.~4, pp. 887--901, 2015.

\bibitem{Ren17_kais}
Y.~Ren, C.~Domeniconi, G.~Zhang, and G.~Yu, ``Weighted-object ensemble
  clustering: methods and analysis,'' \emph{Knowledge and Information Systems},
  vol.~51, no.~2, pp. 661--689, 2017.

\bibitem{Levandowsky1971}
M.~Levandowsky and D.~Winter, ``Distance between sets,'' \emph{Nature}, vol.
  234, pp. 34--35, 1971.

\bibitem{lovasz1993random}
L.~Lov{\'a}sz, ``Random walks on graphs: A survey,'' \emph{Combinatorics, Paul
  Erd\"{o}s is Eighty}, vol.~2, no.~1, pp. 1--46, 1993.

\bibitem{newman04}
M.~E.~J. Newman and M.~Girvan, ``Finding and evaluating community structure in
  networks,'' \emph{Physical Review E}, vol.~69, no. 026113, 2004.

\bibitem{pon05_rw}
P.~Pons and M.~Latapy, ``Computing communities in large networks using random
  walks,'' in \emph{Proc. of International Symposium on Computer and
  Information Sciences (ISCIS)}, 2005, pp. 284--293.

\bibitem{lai_PRE10}
D.~Lai, H.~Lu, and C.~Nardini, ``Enhanced modularity-based community detection
  by random walk network preprocessing,'' \emph{Physical Review E}, vol.~81,
  no. 066118, pp. 264--323, 2010.

\bibitem{tan2005introduction}
P.-N. Tan, M.~Steinbach, and V.~Kumar, ``Introduction to data mining,'' 2005.

\bibitem{lecun98}
Y.~LeCun, L.~Bottou, Y.~Bengio, and P.~Haffner, ``Gradient-based learning
  applied to document recognition,'' \emph{Proceedings of the IEEE}, vol.~86,
  no.~11, pp. 2278--2324, 1998.

\bibitem{Bache+Lichman:2013}
\BIBentryALTinterwordspacing
K.~Bache and M.~Lichman, ``{UCI} machine learning repository,'' 2017. [Online].
  Available: \url{http://archive.ics.uci.edu/ml}
\BIBentrySTDinterwordspacing

\bibitem{vinh2010_ARI}
N.~X. Vinh, J.~Epps, and J.~Bailey, ``Information theoretic measures for
  clusterings comparison: Variants, properties, normalization and correction
  for chance,'' \emph{Journal of Machine Learning Research}, vol.~11, no.~11,
  pp. 2837--2854, 2010.

\end{thebibliography}

\begin{IEEEbiography}[{\includegraphics[width=1in,height=1.25in,clip,keepaspectratio]{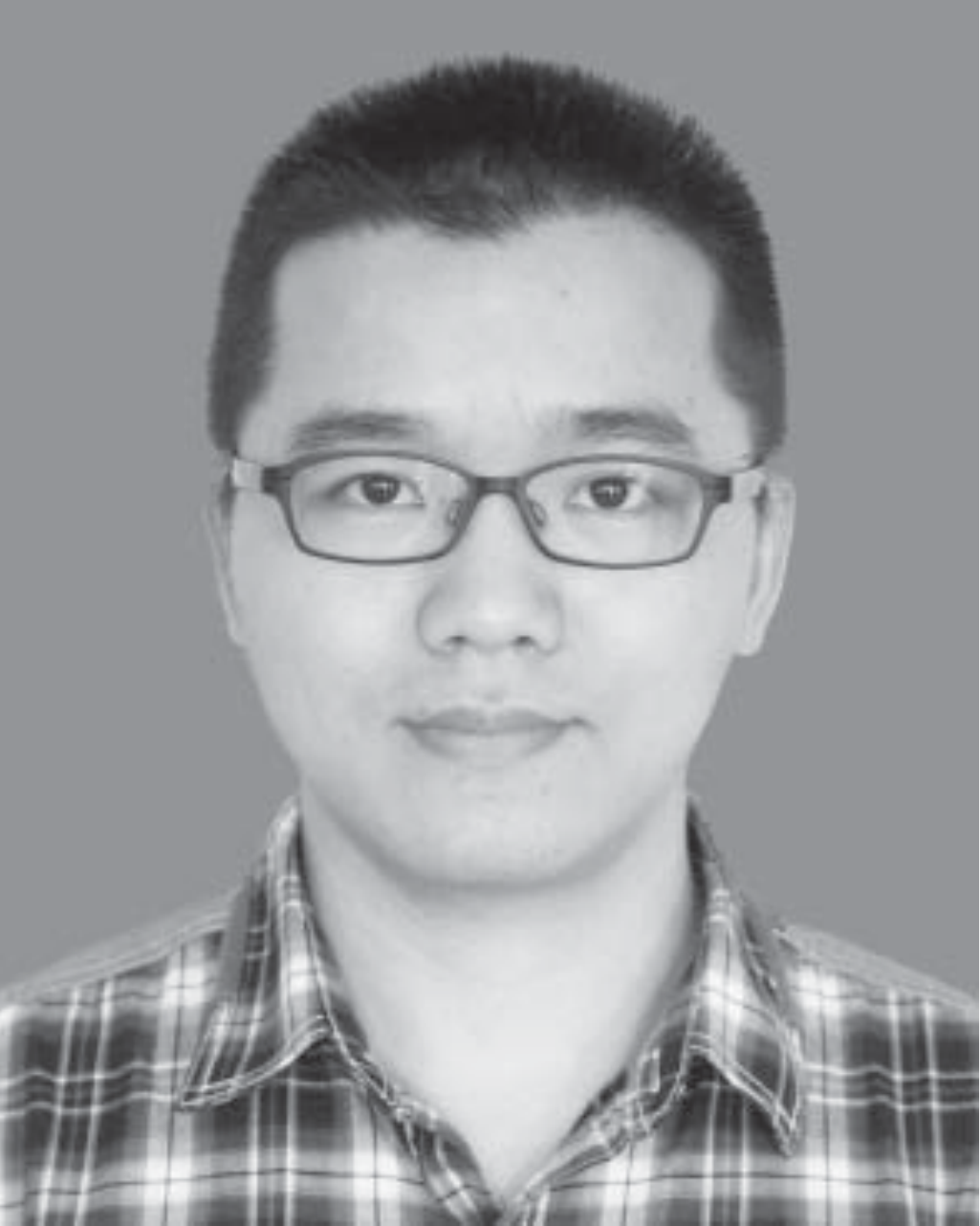}}]{Dong Huang}
received the B.S. degree in computer science in 2009 from South China University of Technology, Guangzhou, China. He received the M.Sc. degree in computer science in 2011 and the Ph.D. degree in computer science in 2015, both from Sun Yat-sen University, Guangzhou, China. He joined South China Agricultural University in 2015, where he is currently an Associate Professor with the College of Mathematics and Informatics. From July 2017 to July 2018, he was a visiting fellow with the School of Computer Science and Engineering, Nanyang Technological University, Singapore. His research interests include data mining and pattern recognition.
\end{IEEEbiography}

\begin{IEEEbiography}[{\includegraphics[width=1in,height=1.25in,clip,keepaspectratio]{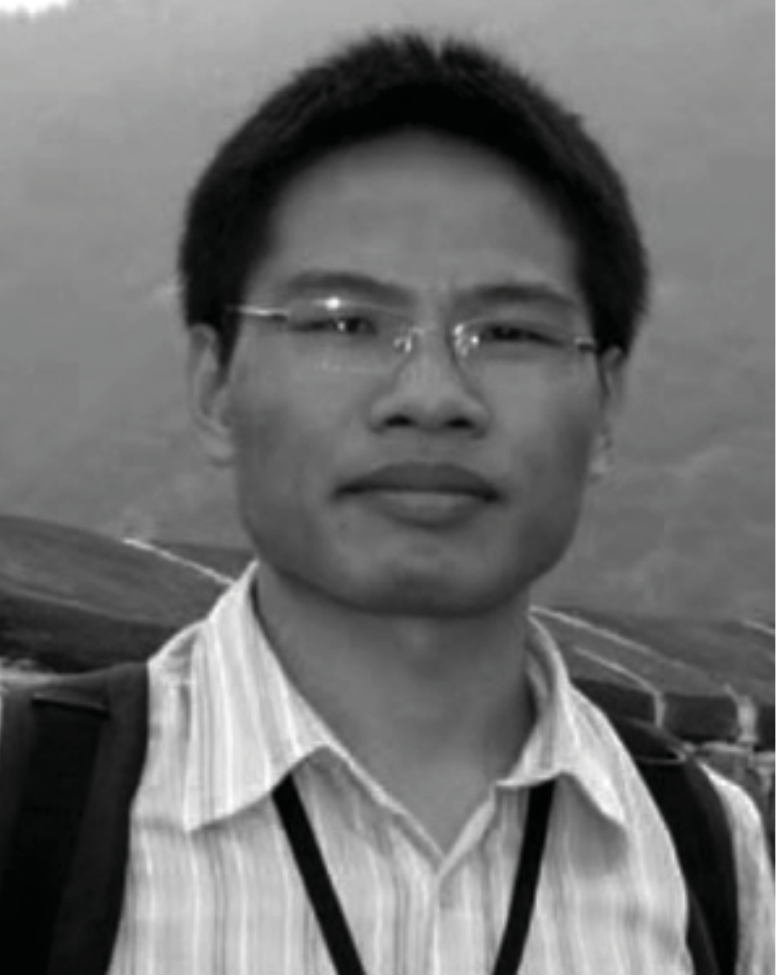}}]{Chang-Dong Wang}
received the Ph.D. degree in computer
science in 2013 from Sun Yat-sen University,
China. He is a visiting student at University of Illinois at Chicago
from Jan. 2012 to Nov. 2012. He joined Sun Yat-sen University in
2013 as an assistant professor and now he is currently an associate professor with School of Data and Computer Science.
His research interests include machine learning
and data mining. He has published over 90 papers in
international journals and conferences such as IEEE TPAMI, IEEE
TKDE, IEEE TCYB, IEEE TSMC-C, Pattern Recognition, KAIS, ICDM, CIKM and SDM. His ICDM 2010 paper won the Honorable Mention for Best
Research Paper Awards. He won 2012 Microsoft Research Fellowship Nomination Award. He was awarded 2015 Chinese Association for Artificial Intelligence (CAAI) Outstanding Dissertation.
\end{IEEEbiography}

\begin{IEEEbiography}[{\includegraphics[width=1in,height=1.25in,clip,keepaspectratio]{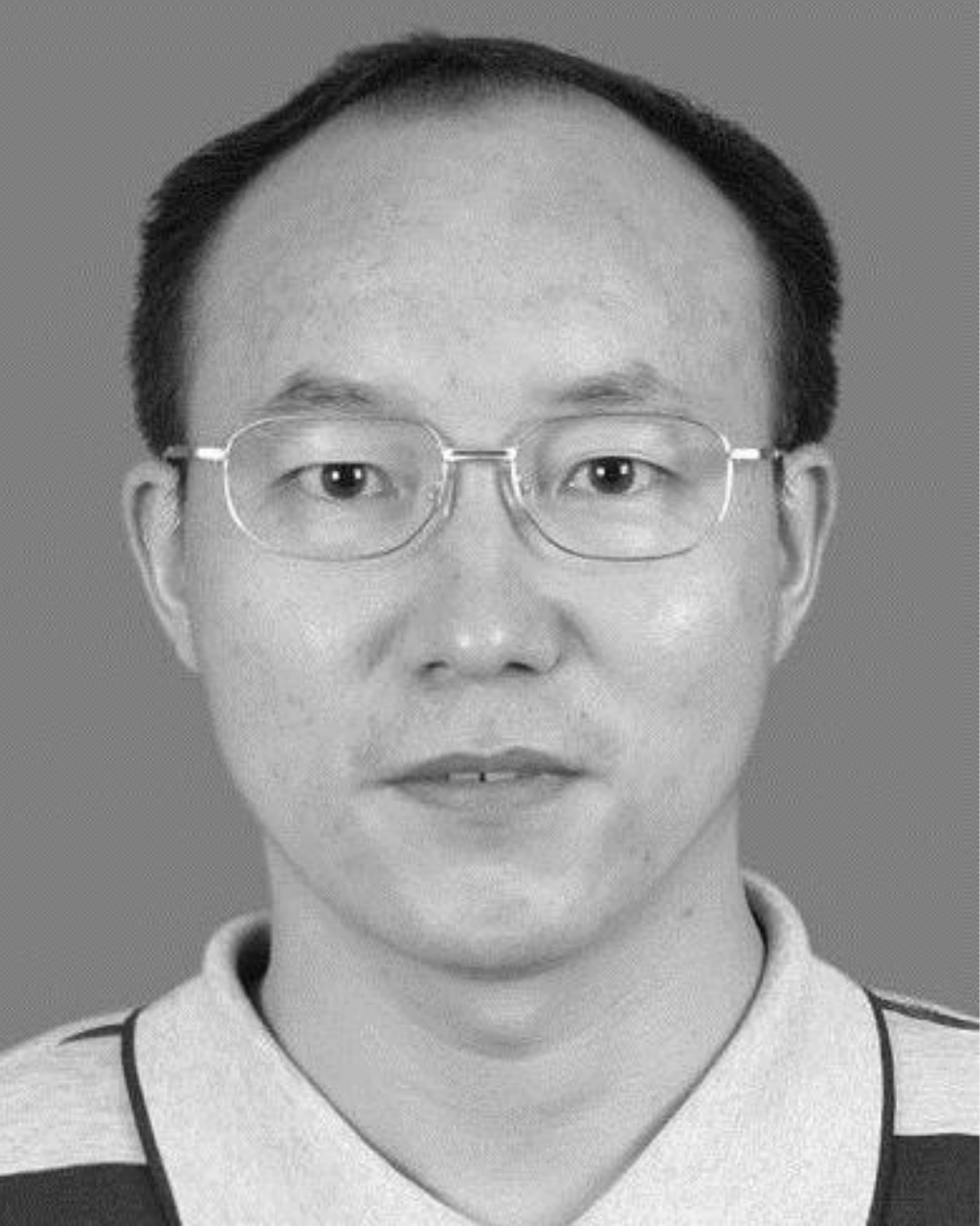}}]{Hongxing Peng}
received his Ph.D. degree in 2014 from South China Agricultural University, Guangzhou, China, where he is currently an Associate Professor with the College of Mathematics and Informatics. From Feb. 2016 to Feb. 2017, he worked as a postdoctoral researcher in Washington State University, Pullman, Washington, U.S. His current research interests include pattern recognition and computer vision.
\end{IEEEbiography}

\begin{IEEEbiography}[{\includegraphics[width=1in,height=1.25in,clip,keepaspectratio]{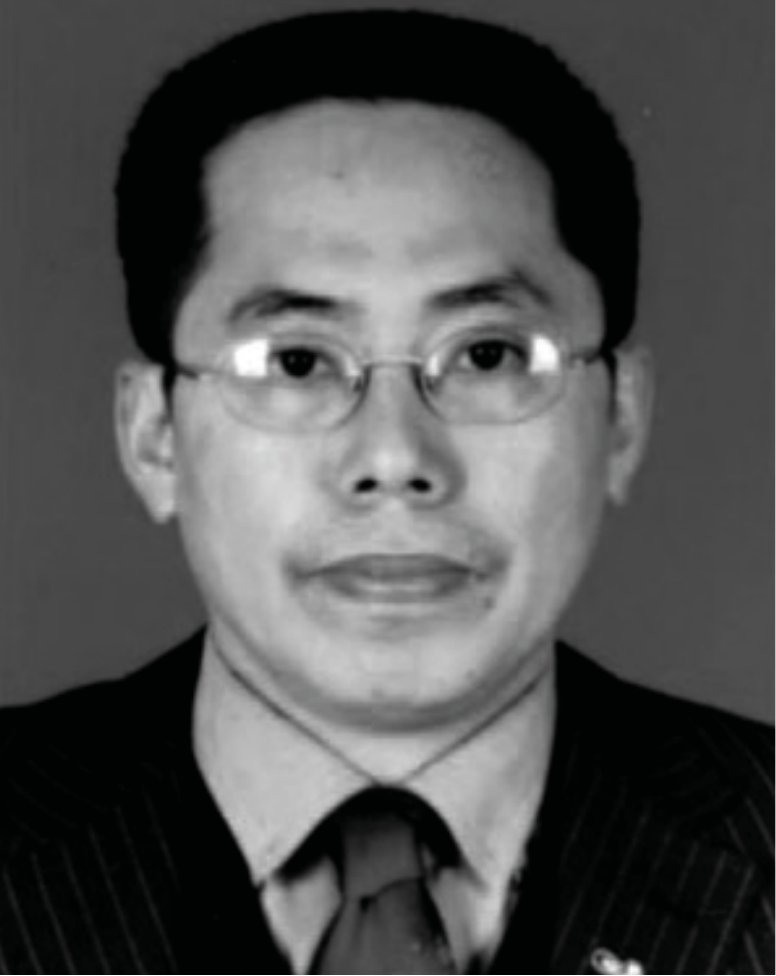}}]{Jianhuang Lai}
received the M.Sc. degree in applied mathematics in 1989 and the Ph.D. degree in mathematics in 1999 from Sun Yat-sen University, China. He joined Sun Yat-sen University in 1989 as an Assistant Professor, where he is currently a Professor with the School of Data and Computer Science. His current research interests include the areas of digital image processing, pattern recognition, multimedia communication, wavelet and its applications. He has published more than 200 scientific papers in the international journals and conferences on image processing and pattern recognition, such as IEEE TPAMI, IEEE TKDE, IEEE TNN, IEEE TIP, IEEE TSMC-B, Pattern Recognition, ICCV, CVPR, IJCAI, ICDM and SDM. Prof. Lai serves as a Standing Member of the Image and Graphics Association of China, and also serves as a Standing Director of the Image and Graphics Association of Guangdong.
\end{IEEEbiography}

\begin{IEEEbiography}[{\includegraphics[width=1in,height=1.25in,clip,keepaspectratio]{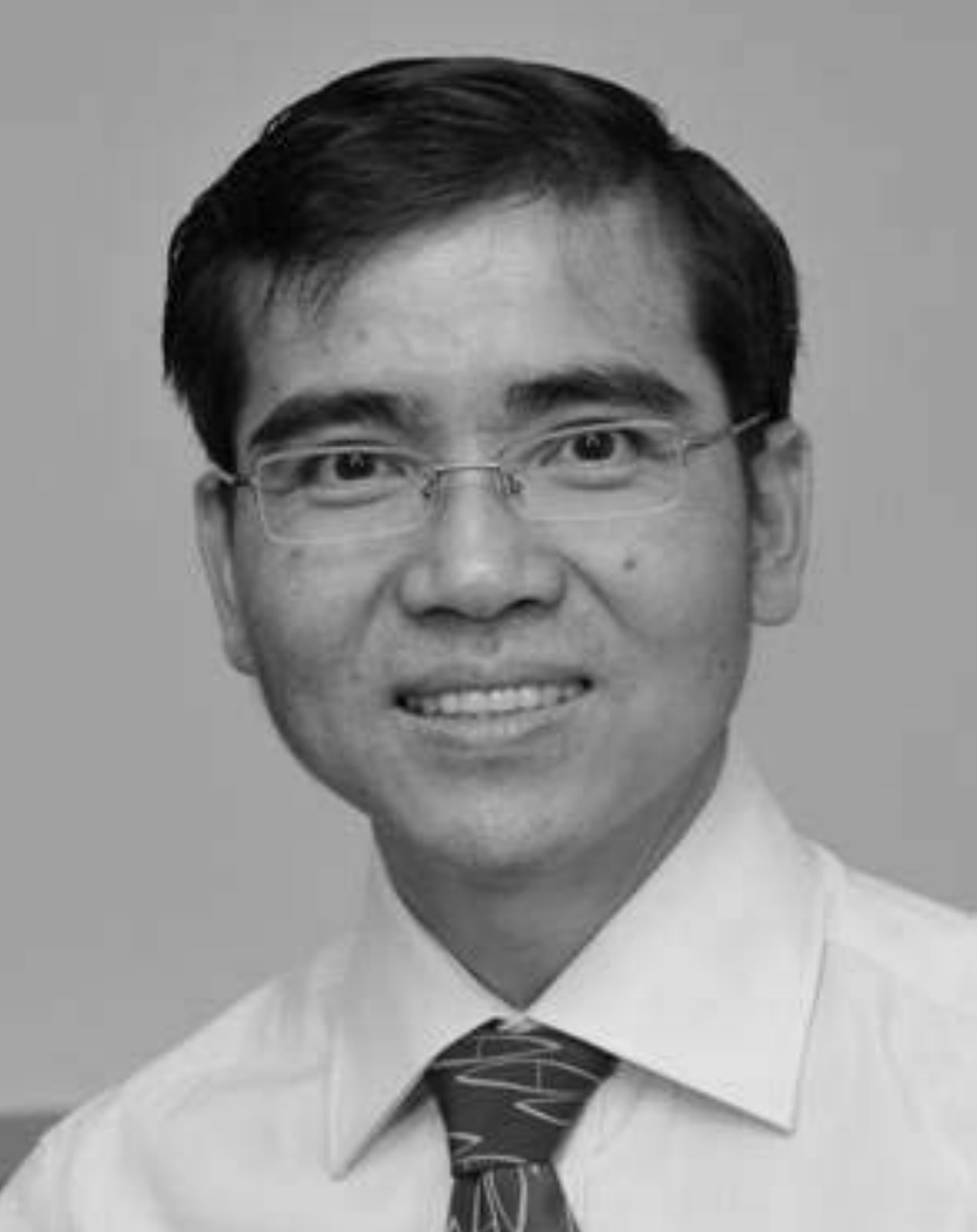}}]{Chee-Keong Kwoh} received the bachelor's degree in electrical engineering (first class) and the master's degree in industrial system engineering from the National University of Singapore in 1987 and 1991, respectively. He received the PhD degree from the Imperial College of Science, Technology and Medicine, University of London, in 1995. He has been with the School of Computer Science and Engineering, Nanyang Technological University (NTU) since 1993. He is the programme director of the MSc in Bioinformatics programme at NTU. His research interests include data mining, soft computing and graph-based inference; applications areas include bioinformatics and biomedical engineering. He has done significant research work in his research areas and has published many quality international conferences and journal papers. He is an editorial board member of the International Journal of Data Mining and Bioinformatics; the Scientific World Journal; Network Modeling and Analysis in Health Informatics and Bioinformatics; Theoretical Biology Insights; and Bioinformation. He has been a guest editor for many journals such as Journal of Mechanics in Medicine and Biology, the International Journal on Biomedical and Pharmaceutical Engineering and others. He has often been invited as an organizing member or referee and reviewer for a number of premier conferences and journals including GIW, IEEE BIBM, RECOMB, PRIB, BIBM, ICDM and iCBBE. He is a member of the Association for Medical and Bio-Informatics, Imperial College Alumni Association of Singapore. He has provided many services to professional bodies in Singapore and was conferred the Public Service Medal by the President of Singapore in 2008.
\end{IEEEbiography}

\end{document}